\documentclass[journal]{IEEEtran}

\usepackage{xcolor,soul,framed} 

\colorlet{shadecolor}{yellow}
\usepackage[pdftex]{graphicx}
\graphicspath{{../pdf/}{../jpeg/}}
\DeclareGraphicsExtensions{.pdf,.jpeg,.png}

\usepackage[cmex10]{amsmath}
\usepackage{array}
\usepackage{mdwmath}
\usepackage{mdwtab}
\usepackage{eqparbox}
\usepackage{url}
\usepackage{algorithm}
\usepackage{algorithmic}
\usepackage{graphicx}
\usepackage{subfigure}
\usepackage{booktabs}
\usepackage{amsthm}

\begin{document}
\bstctlcite{IEEEexample:BSTcontrol}
    \title{Fine-grained Graph Learning for Multi-view Subspace Clustering}
  \author{Yidi~Wang,
          Xiaobing~Pei,
          Haoxi~Zhan\\

  \thanks{Manuscript received 9 December 2022; revised 2 April 2023; accepted 21 June 2023. \emph{(Corresponding author: Xiaobing~Pei.)}}
  \thanks{Yidi Wang, Xiaobing Pei are with the School of Software, Huazhong University of Science and Technology, Wuhan 430074, China (email: m202076655@hust.edu.cn; xiaobingp@hust.edu.cn).}
  \thanks{Haoxi Zhan is with the Faculty of Electronic Information and Electrical Engineering, Dalian University of Technology, Dalian 116000, China (email: zhanhaoxi@foxmail.com).}
  \thanks{© 2023 IEEE.  Personal use of this material is permitted.  Permission from IEEE must be obtained for all other uses, in any current or future media, including reprinting/republishing this material for advertising or promotional purposes, creating new collective works, for resale or redistribution to servers or lists, or reuse of any copyrighted component of this work in other works.}
}

\markboth{PREPRINT. ACCEPTED IN: IEEE TRANSACTIONS ON EMERGING TOPICS IN COMPUTATIONAL INTELLIGENCE, JUNE 2023
}{Wang \MakeLowercase{\textit{et al.}}: FINE-GRAINED GRAPH LEARNING FOR MULTI-VIEW SUBSPACE CLUSTERING}

\maketitle

\begin{abstract}
Multi-view subspace clustering (MSC) is a popular unsupervised method by integrating heterogeneous information to reveal the intrinsic clustering structure hidden across views. Usually, MSC methods use graphs (or affinity matrices) fusion to learn a common structure, and further apply graph-based approaches to clustering. Despite progress, most of the methods do not establish the connection between graph learning and clustering. Meanwhile, conventional graph fusion strategies assign coarse-grained weights to combine multi-graph, ignoring the importance of local structure. 
In this paper, we propose a fine-grained graph learning framework for multi-view subspace clustering (FGL-MSC) to address these issues. To utilize the multi-view information sufficiently, we design a specific graph learning method by introducing graph regularization and a local structure fusion pattern.
The main challenge is how to optimize the fine-grained fusion weights while generating the learned graph that fits the clustering task, thus making the clustering representation meaningful and competitive. Accordingly, an iterative algorithm is proposed to solve the above joint optimization problem, which obtains the learned graph, the clustering representation, and the fusion weights simultaneously. Extensive experiments on eight real-world datasets show that the proposed framework has comparable performance to the state-of-the-art methods.
The source code of the proposed method is available at https://github.com/siriuslay/FGL-MSC.

\end{abstract}


%
\IEEEpeerreviewmaketitle


\section{Introduction}

\IEEEPARstart{A}{S} an effective unsupervised learning method, multi-view clustering has attracted wide attention in recent years \cite{msc-1, msc-2}. Among this literature, the graph-based clustering methods are brought forward and have achieved remarkable improvements in a series of applications. Graph-based approaches have been widely adopted in the early period of single-view analysis. The process can be interpreted as  a two-stage algorithm: constructing graphs from features and using graph Laplacian to solve a quadratic optimization problem \cite{SC1}. These methods can generally converge to a global optimal solution and identify arbitrary shapes of clusters. Despite progress, graph-based clustering methods are often limited by the quality of the generated graphs. In this regard, many graph construction methods are proposed. The classic ones are the fully-connected graph constructed by the Gaussian kernel function and the k-nearest neighbor graph which only retains part of edges \cite{KNN}. In addition, there are local discriminant graph \cite{LDG}, pairwise similarity graph \cite{PSG}, adaptive neighbors graph \cite{CAN}, and so on.

A different graph construction approach based on subspace learning is proposed by \cite{subspace}, and it further developed a new subspace clustering pattern e.g. \cite{subspace1,subspace2,subspace3,LMSC,GLMSC}. Subspace clustering methods can capture more global manifold information by using a self-expression matrix, which can deal with noise and missing points \cite{sub2}. With the widespread use of neural networks, more works have extended subspace clustering to deep learning and achieved better results than traditional models  e.g. \cite{deep1,sub1,sub3,deep2,deep3}.

Through the in-depth study of clustering, more and more researchers find that single-view data cannot completely describe the internal structure of data. Therefore, multi-view clustering methods are gradually developed. \cite{Consistency} and \cite{CSMSC} suggest that the multi-view clustering models are generally based on consistency across views. \cite{mvc1} further concludes that the consensus principle and the complementary principle play important roles in the success of multi-view clustering. Specifically, the consensus principle means that each view admits a common clustering structure and the complementary principle means that each view contains different degrees of variation in the distribution of samples. 
Hence the approaches of learning one potential common subspace or graph structure and further deriving the cluster indicator are feasible by integrating multi-view information. 

Practically, there are three prevailing schemes to cope with the integration of multi-view information. The first kind tries to strike a balance between consistency and inconsistency, by assigning weights for views to fuse rich information, such as the multi-view subspace clustering model \cite{MVSC}, auto-weighted multiple graph learning model \cite{AMGL}, multi-view graph learning model \cite{MVGL}, multi-view non-negative embedding and spectral embedding model \cite{NESE}, and its constraint form \cite{CNESE}. The second kind is to decompose the original graph structure captured on a single view into two parts, that is, a common consensus graph and an independent graph, and optimize to obtain the graph structures and fusion weights of both parts, such as consistent and specific multi-view subspace clustering model \cite{CSMSC}, a consistency-induced multi-view subspace clustering model \cite{CiMSC}, and a unified graph alignment framework \cite{VCGA}. In addition, many tensor-based analytical methods have emerged in recent years, mostly by direct data mining on tensors composed of multi-view data, exploiting tensor decomposition and multi-rank constraints to filter redundant information and search for potential common structure, e.g. \cite{tensor1,ETLMSC,tensor2,LTBPL,tensor4,tensor6}. By nature, tensor-based methods also focus on the retention of consistent information, but the data processing and solution forms differ from the above two kinds.

\begin{figure}
\centering
\vspace{0in}
\subfigtopskip=0pt 
\subfigcapskip=-5pt 
\subfigure[Neighbourhood Level]{
	\label{mistake} 
	\includegraphics[width=0.45\linewidth]{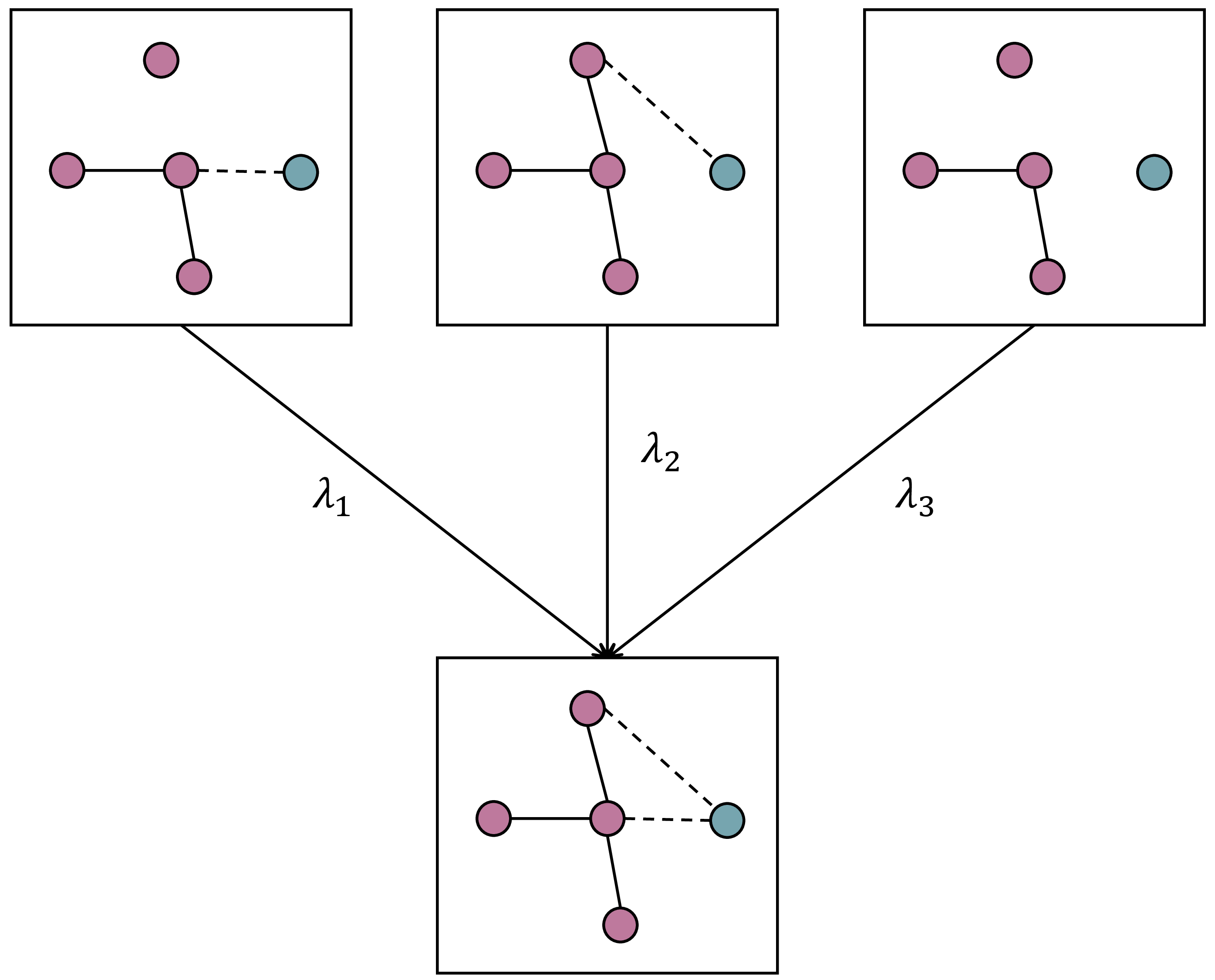}}
\hspace{0.1in}
\subfigure[Cluster Level]{
	\label{redundant}
	\includegraphics[width=0.45\linewidth]{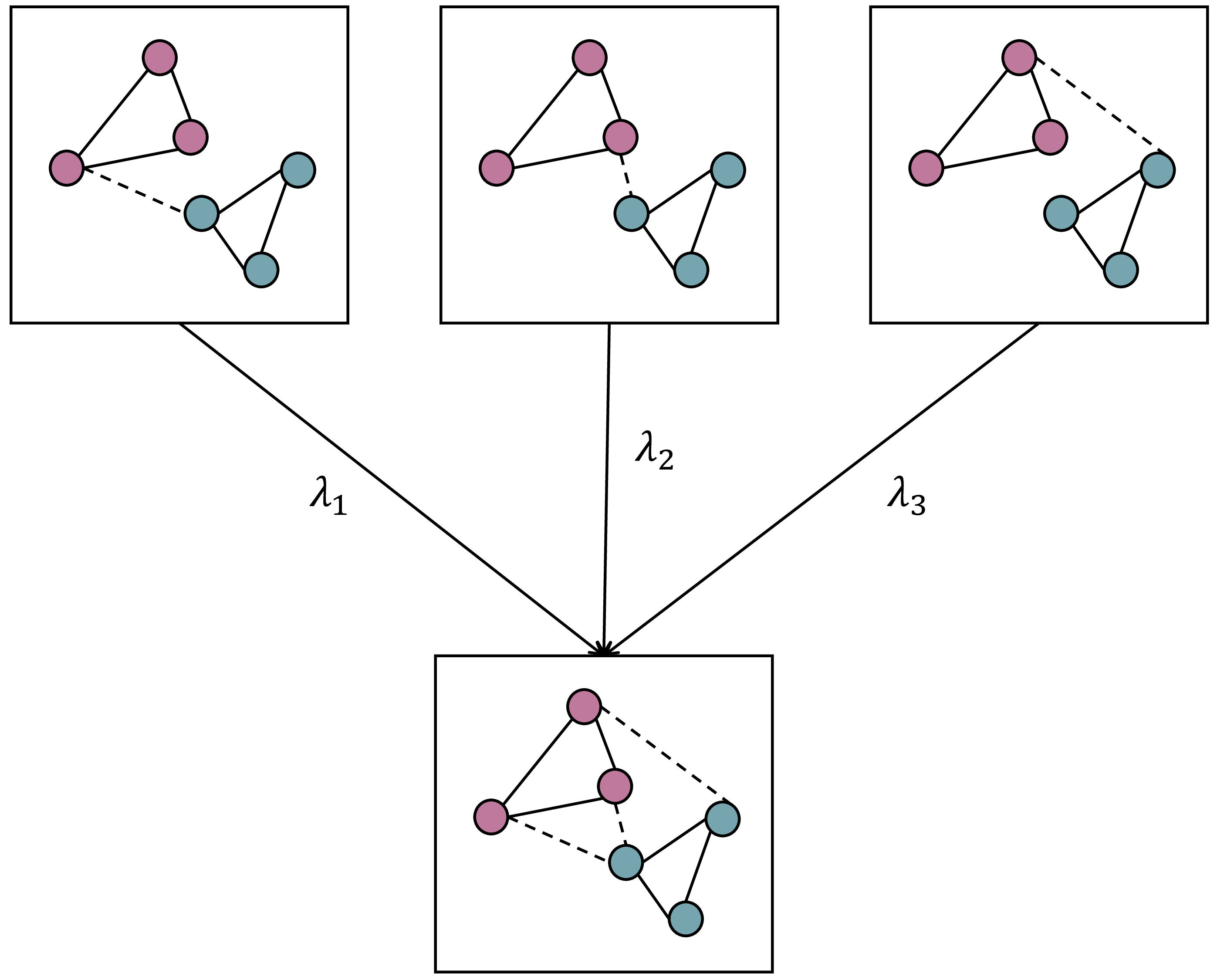}}
\caption{Two types of problems in multi-view clustering. Given two colors of points representing two clusters, the solid and dashed lines represent desired and redundant edges, respectively. In Fig.1(a), the wrong neighbors will be remembered caused by the coarse-grained weights $\lambda$. In Fig.1(b), the redundant structure between two clusters will not be cleared up due to the linear combination.}
\label{problem} 
\end{figure}
Although the aforementioned methods of multi-view graph learning and clustering have achieved excellent performance, most of them still suffer from problems in graph structure fusion and weight learning. For example, if there are local structure mistakes in several views that result in a gap between the learned graph and the ideal clustering structure, the coarse-grained weights will perpetuate these mistakes. The other situation is that all views have parts of redundant structures that do not conform to the real clustering structure. It is difficult for the linear combinations at the view level to remove redundancy. On the contrary, it even leads to the superimposition of redundancies and thus acquires a more imprecise clustering structure. Fig.\ref{problem} shows the above two problems clearly. To overcome the above challenges, \cite{MVGL} proposes a novel fine-grained graph fusion pattern to process the intermediate graphs more delicately. But its framework divides into two parts, which graph learning and clustering conduct separately, resulting in a compromised clustering performance. Some investigations verify that a unified two-part framework can effectively improve model performance e.g. \cite{VCGA,MVC-NMF,AwSCGLD}. Therefore, more ingenious strategies for multi-view subspace clustering are urgently needed.

\begin{figure*}
  \begin{center}
  \includegraphics[width=\linewidth]{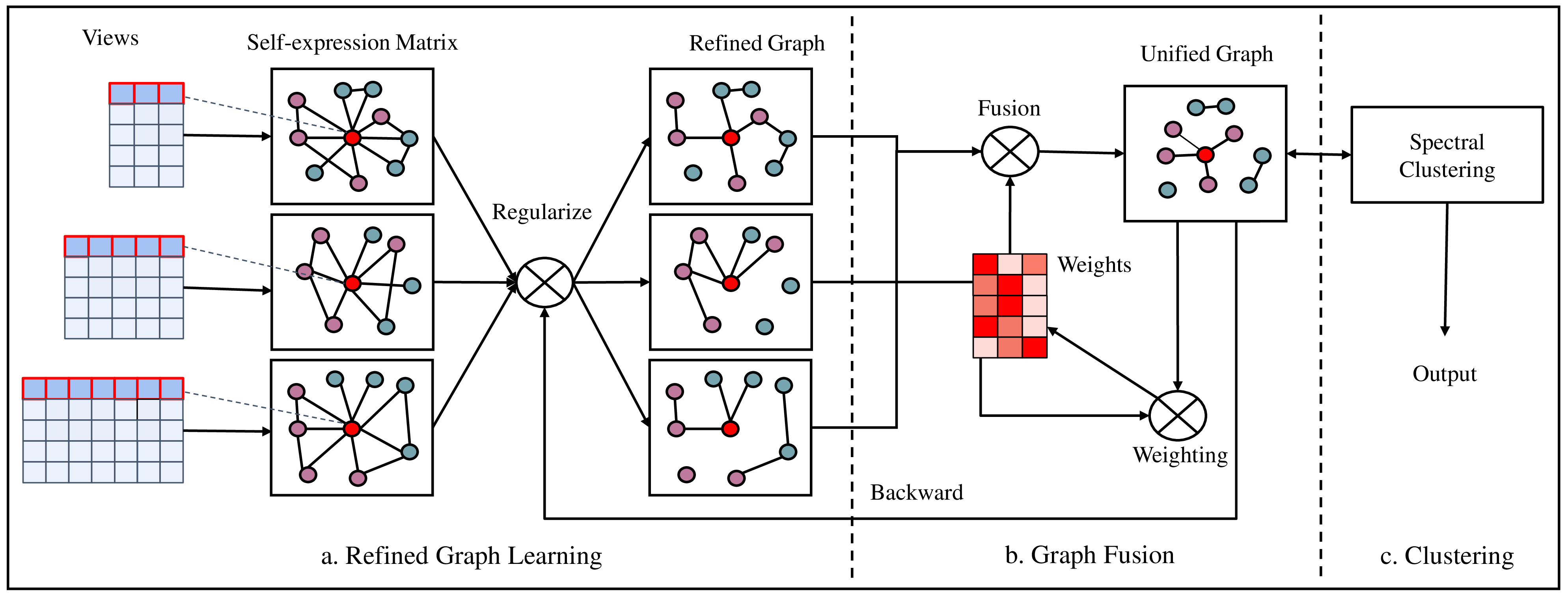}\\
  \caption{An illustration to our proposed FGL-MSC.}\label{model}
  \end{center}
\end{figure*}

In this paper, we propose a novel multi-view subspace clustering framework, denoted by Fine-grained Graph Learning for Multi-view Subspace Clustering (FGL-MSC). As is shown in Fig.\ref{model}, FGL-MSC contains three important parts: the refined graphs generated by the self-expression matrix and graph regularization, 
fine-grained graph fusion term, and the reformulation of spectral clustering (the rank constraint term). Compared with the existing methods, our contributions are listed as follows:
\begin{enumerate}
    \item we propose a joint learning framework FGL-MSC, which implements subspace learning, fine-grained weights learning, unified graph learning, and clustering simultaneously. Different from most of the methods in the literature, our unified graph learning strategy takes the original subspace learning and clustering representation into account together. 
    \item We regard the unified graph learning process as an aggregation of local structures for each sample, which considers fine-grained weights to all samples for different views, alleviating the cross-talk of local structures. For the unified graph, a Frobenius norm is also introduced to control its sparsity. 
    \item We design an effective algorithm to solve the optimization of FGL-MSC, and theoretically show its convergence and computational complexity. The experiments on eight real-world datasets show that our method has comparable performance with the SOTA methods.
\end{enumerate}

The rest of this paper is structured as follows. Section II provides the notations and some related works. Section III presents the model formulation and its optimization algorithm. The experiments on benchmark datasets are shown in Section IV. And Section V concludes the paper.

\section{Notations and Related Works}

In this section, we first define the notations used throughout the paper and briefly review three representative multi-view clustering models. The functional form of spectral clustering is also provided.

\noindent 
\textbf{Notation:} Given a multi-view data $\mathcal{X}=\{X^1,\cdots, X^t\} \in R^{d_v \times n}$, $t$, $n$, $d_v$ denote the number of views, the number of samples, and the feature dimension of $X^v$, respectively. $\mathcal{W}=\{W^1,\cdots,W^t\}$ is the self-expression matrices generated from different subspaces. The clustering representation (or the input of clustering tasks) is denoted by $F$. 

For an arbitrary matrix $A$, $A_{ij}$ denotes the $ij$-th element and $A_i$ is the $i$-th column of $A$. $A^T$ and $Tr(A)$ means the transpose and the trace of $A$, respectively. $||A||_1$ denotes the $l_1$-norm of $A$. The Frobenius norm of $A$ is $||A||_F = \sqrt{\sum_{i,j}{{\lvert A_{ij} \rvert }^2}}$. The $l_2$-norm of vector $A_i$ is $||A_i||_2$. $A \geq 0$ means all elements in $A$ are non-negative.

In addition, $\boldsymbol{1}$ denotes a column vector whose elements are all $1$. $I$ denotes an identity matrix with the proper size. All matrices are represented in capital letters.

\subsection {Multi-view Subspace Clustering}

Subspace clustering assumes that samples can be linearly represented by other samples lying in the same subspace, so as to mine the correlation between different data. Mathematically, given $X\in R^{d\times n}$ with $n$ samples and $d$ is the feature dimension. The subspace clustering can be formulated as \cite{sub00}:
\begin{equation} \label{SC}
    \min_{W} ||X-X W||^{2}_{F} + \lambda \Omega(W),
\end{equation}
where $W \in R^{n\times n}$ is the self-expression matrix, $\Omega(\cdot)$ is a regularized function, and $\lambda \textgreater 0$ is a balance parameter. The MVSC model \cite{MVSC} extended Eq.(\ref{SC}) to the multi-view problems. The specific form is:
\begin{equation} 
\begin{aligned}\label{MVSC} 
    &\min_{W^v,F} ||X^v-X^v W^v||^{2}_{F} + \lambda \sum_v Tr(F^TL^v F)\\
    &s.t. \quad W^v \geq 0, W^v \boldsymbol{1}=\boldsymbol{1}, F^T F=I,
\end{aligned}
\end{equation}
where $L^v = D^v-W^v$ is the graph Laplacian matrix, and $D^v$ is the degree matrix with $D^v_{ii}=\sum^n_{j=1} W^v_{ij}$. Eq.(\ref{MVSC}) learns the self-expression matrix for each view, and then obtains the common clustering representation directly by the rank constraint term. The information integration of this method is reflected in the optimization of clustering representation $F$.

\subsection{Multi-graph Fusion for Multi-view Spectral Clustering}

GFSC \cite{GFSC} proposes a new graph fusion mechanism to integrate different subspaces into one graph, and optimize clustering representation together:
\begin{equation} 
\begin{aligned}\label{GFSC} 
    &\min_{W^v,S,F} \sum^t_{v=1} ( ||X^v-X^v W^v||^2_F + \alpha||W^v||^2_F \\
    &\qquad + \beta \lambda_v||W^v-S||^2_F ) + \gamma Tr(F^T L_S F)\\
    &s.t. \quad W^v \geq 0, F^T F=I,
\end{aligned}
\end{equation}
where $S$ is a consensus graph, which approximates the self-expression matrix of each individual view and maintains an explicit clustering structure via the spectral clustering term. $L_S$ is the Laplacian matrix of $S$, with the same definition as Eq.(\ref{MVSC}). The graph fusion weight $\lambda_v$ is determined by the inverse distance between each individual graph $W^v$ and the consensus graph $S$. When the $\lambda$ is set to $\frac{1}{t}\boldsymbol{1}$, the GFSC model degrades to the MVSC model.

\subsection{Graph Learning for Multi-view Clustering}

Different from the above two integrating methods, the MVGL model \cite{MVGL} proposes a node-level graph learning method. The key in MVGL is a two-step graph learning process. Firstly, it calculates a set of single view graphs by adaptive neighbors graph \cite{CAN} as follows:
\begin{equation}\label{singlegraph} 
\begin{aligned}
    &\min_{W^v} \sum_{i,j=1}^n W_{ij}^v ||X_i^v-X_j^v||_2^2 + \lambda ||W^v||_F^2\\
    &s.t. \quad W^v \geq 0, W_i^T \boldsymbol{1}=1.
\end{aligned}
\end{equation}

Then assuming that $G$ is the unified graph fused by all the $W^v$, it proposes a global graph learning function:

\begin{equation} \label{globalgraph} 
\begin{aligned} 
    &\min_{G, \alpha} \sum_{i=1}^n {||G_i - \sum_{v=1}^t {\alpha_i^v W_i^v}||_2^2} + \gamma Tr(F^T L_G F)\\
    &s.t. \quad \forall{i}, \sum_{v=1}^t {\alpha_i^v} = 1, {\alpha_i \geq 0},\\
    &\qquad G \geq 0, G\boldsymbol{1}=\boldsymbol{1}, F^T F=I,
\end{aligned}
\end{equation}
where $ \alpha $ is the fine-grained fusion weights. and $L_G$ denotes the Laplacian matrix of $G$. Based on Eq.(\ref{globalgraph}), MVGL turns the local structure fusion into a least square approximation problem, and provides an efficient optimization algorithm.

\subsection{Reformulation of Spectral Clustering}

The aforementioned methods use the reformulation of spectral clustering to learn the graph and clustering representation simultaneously. Here we make a brief description of this popular technology.

Let $c$ denote the cluster number and $F \in R^{n\times c}$ be the clustering representation vector. Given a graph Laplacian matrix $L=D-W$, where $W$ is the generated graph from data and $D$ is a degree matrix of $W$, \cite{AMGL} provides the functional form of spectral clustering to obtain $F$:
\begin{equation} \label{spectral}
    \min Tr(F^T L F) \quad s.t.\quad F^T F=I.
\end{equation}

The optimal solution $F$ to the Eq.(\ref{spectral}) is formed by the $c$ eigenvectors of $L$ corresponding to the $c$ smallest eigenvalues. Furthermore, given $F$, the construction of an optimal graph with $c$ connected components is also equivalent to Eq.(\ref{spectral}) according to following equation \cite{CAN}:
\begin{equation} \label{spectral2}
    \sum_{i=1}^c\sigma_i(L)=\min_{F^T F=I} Tr(F^T L F),
\end{equation}
where $\sigma_i(L)$ is the $i$-th smallest eigenvalue of $L$. Therefore, Eq.(\ref{spectral}) is introduced into many multi-view clustering frameworks to participate in joint optimization as a rank constraint of graphs.

\section{Methodology}

In this section, we first derive the proposed model formulation, then present its optimization algorithm, and finally provide the overall analysis of the model.

\subsection{Model Formulation}

As shown in Fig.(\ref{model}), FGL-MSC can be divided into three stages.

\subsubsection{Refined Graph Learning}

As aforementioned, the self-expression matrix $W^v$ obtained by Eq.(\ref{SC}) can capture global information but ignores the local structure. In recent works, the distance metric graphs are frequently used to complement the self-expression matrix in the description of the local structure \cite{LEnewS, MVLEnewS, SGSK}. The processed (or refined) graph shows stronger robustness than the original self-expression matrix. Inspired by these investigations, we design a two-part learner to extract useful structure from $\mathcal{W}$. The learner's form can be described as :
\begin{equation}\label{Graphlearner} 
\begin{aligned}
    &\min \Omega_1(W^v,Z^v) + \lambda \Omega_2(G,Z^v,\alpha_v) \\
    &s.t. \quad W^v \geq 0, Z^v \geq 0, G \geq 0,
\end{aligned}
\end{equation}
where $G$ is the unified graph, $\alpha_v$ is fusion weights, and $Z^v$ denotes the refined graph. $\lambda \textgreater 0$ is a hyper-parameter.

In the first part of Eq.(\ref{Graphlearner}), we design a graph regularization term to ensure the retention of global information. Meanwhile, $Z^v$ is initialized by the adaptive neighbors graph \cite{CAN}, which extracts the local structure efficiently. In the same way as Eq.(\ref{singlegraph}), The initialization of $Z^v$ can be described as follows:

\begin{equation}\label{CAN} 
\begin{aligned}
    &\min_{Z^v} \sum_{i,j=1}^t Z_{ij}^v ||X_i^v-X_j^v||_F^2 + \gamma ||Z^v||_F^2\\
    &s.t. \quad Z^v \geq 0, \forall{i}, Z_i^T \boldsymbol{1}=1.
\end{aligned}
\end{equation}

\subsubsection{Graph Fusion}

The second part of Eq.(\ref{Graphlearner}) is a graph fusion step. Generally, $\alpha$ to each view is often optimized independently, which depends a lot on the overall quality of its view. To address the rough weights assignment caused by this dependence, we adopt the fine-grained fusion weights proposed by \cite{MVGL}.

To overcome the inconvenience of vector representation, We transform the tensor $\mathcal{Z}$ at the data level:
\begin{equation}\label{trans}
    \widetilde{\mathcal{Z}}=trans(\mathcal{Z}),
\end{equation}
where the transformation function $trans(\mathcal{Z})$ can be described as $\widetilde{Z}^i = [(Z_i^1),\cdots,(Z_i^t)]^T$. The obtained $\widetilde{\mathcal{Z}}=\{\widetilde{Z}^1,\cdots,\widetilde{Z}^n\} \in R^{t \times n}$ denotes multi-view topology for all samples, which can be considered as a cross-view structural representation. Therefore, the specific formulation of Eq.(\ref{Graphlearner}) is as follows:
\begin{equation} \label{FGL2}
\begin{aligned} 
    &\min_{G, A} \sum_{v=1}^t ||W^v-Z^v||_F^2 + \lambda \sum_{i=1}^n ||G_i^T - A_i^T \widetilde{Z}^i||^2_2\\
    &s.t. \quad \forall{i}, A_i^T \boldsymbol{1}=1, A_i \geq 0, G \geq 0, G\boldsymbol{1}=\boldsymbol{1},
\end{aligned}
\end{equation}
where $A_{iv}$ is the weight for the $i$-th sample in the $v$-th view. To ensure that the weights are meaningful, we add a constraint that the sum of each column of $A$ is one. $A_i$ is optimized independently. $\lambda \textgreater 0$ is a trade-off hyper-parameter. This problem is a classical convex quadratic programming (QP), which can be solved by some QP optimizers. Our specific optimization algorithm will be described in detail later.

\subsubsection{The Unified Model}

Furthermore, taking advantage of the functional spectral clustering, we establish the connection between graph learning and clustering. The unified objective function combining two tasks can optimize them simultaneously. Considering Eq.(\ref{MVSC}) and Eq.(\ref{FGL2}), the joint optimization problem is as follows:
\begin{equation} \label{FGL3}
\begin{aligned}
    &\min_{G,F,A} \sum_{v=1}^t \{||X^v-X^vW^v||^{2}_{F} + \alpha||W^v-Z^v||_F^2 \} \\
    & \qquad + \lambda \sum_{i=1}^n ||G_i^T - A_i^T \widetilde{Z}^i||^2_2 + \eta Tr(F^T L F)\\
    &s.t. \quad \forall{i}, A_i^T \boldsymbol{1}=1, A_i \geq 0, G \geq 0, G\boldsymbol{1}=\boldsymbol{1}, F^T F=I,
\end{aligned}
\end{equation}
where $\alpha \textgreater 0, \lambda \textgreater 0, \eta \textgreater 0$ are three trade-off hyper-parameters, and $L$ is the Laplacian matrix of $G$. Note that Eq.(\ref{FGL3}) is prone to yielding the trivial solution with respect to $G$, i.e., all non-zero elements are located on the diagonal of $G$, while $L = 0$. So we use the Frobenius norm of $G$ to smooth its elements.

Then our Fine-grained graph Learning for Multi-view Subspace Clustering (FGL-MSC) can be formulated as:
\begin{equation} \label{FGL-MSC}
\begin{aligned}
    &\min_{G,F,A} \sum_{v=1}^t \{||X^v-X^v W^v||^{2}_{F} + \alpha||W^v-Z^v||_F^2 \\
    & \qquad + ||W^v||_1\} + \lambda \sum_{i=1}^n ||G_i^T - A_i^T \widetilde{Z}^i||^2_2 +\gamma ||G||_F^2\\
    & \qquad + \eta Tr(F^T L F)\\
    &s.t. \quad \forall{i}, A_i^T \boldsymbol{1}=1, A_i \geq 0, G \geq 0, G\boldsymbol{1}=\boldsymbol{1}, F^T F=I,
\end{aligned}
\end{equation}
where $\gamma \textgreater 0$ is also a hyper-parameter for the regularization of $G$. Note that $\gamma$ is difficult to be tuned, so we propose an effective algorithm to optimize $G$, which is shown in the next subsection.

This model enjoys the following properties:
\begin{enumerate}
    \item According to Eq.(\ref{trans}), The graph transformation operation transforms generalized sparse graphs into representations of nodes that contain high-level structural information. Then the point-level graph learning term reconstructs them as one sparse graph. Essentially, it's a dimensional reduction from $n \times n \times t$ to $n \times n$, which is a different opinion from existing methods.
    \item The fine-grained weights, i.e., $A$ in Eq.(\ref{FGL2}) can be seen as the reconstruction coefficient during the dimensional reduction. The constraint on this coefficient can ensure the structural feature selection and extraction capability so that the learned structural representation not only retains the latent consistency but also reduces the noise.
    \item The joint framework Eq.(\ref{FGL-MSC}) fulfills the tasks of subspace capturing, graph learning, clustering, and point-level weights learning in a single objective function, which enhances the connection between different tasks.
\end{enumerate}

\subsection{Model Optimization}

We introduce an alternating iteration algorithm to optimize the problem (\ref{FGL-MSC}). We decompose it as five subproblems and update them alternatively.

\subsubsection{$\boldsymbol{W^v}$-subproblem}

When $Z^v,G,A,F$ are fixed, we can obtain the subproblem of Eq.(\ref{FGL-MSC}) with respect to $W^v$:
\begin{equation} \label{Wp}
\begin{aligned}
    &\min_{W^v} ||X^v-X^vW^v||^{2}_{F} + \alpha||W^v-Z^v||_F^2 + ||W^v||_1\\
    &s.t.\quad W^v \geq 0.
\end{aligned}
\end{equation}

It is obvious that Eq.(\ref{Wp}) can not be derived directly due to the non-differentiable property of the Lasso regularization. But there are still many ways to solve it, such as the Proximal Gradient Algorithm (PGA). \cite{LEnewS} investigates a lot in solving the same kind of problems and proposes an efficient algorithm under non-negative data, which has achieved better results. Similarly, we can solve the Eq.(\ref{Wp}) as follows:
\begin{equation} \label{solveW}
    (W^v_{ij})^{k+1}= (W^v_{ij})^k\frac{2[(X^v)^TX^v+\alpha Z^v]_{ij}}{(P_{ij}^v)^k+1},
\end{equation}
where $(P^v)^k=(W^v)^k(X^v)^TX^v+(X^v)^TX^v(W^v)^k+2\alpha (W^v)^k$. The optimal $W^v$ can be easily obtained through several iterations.

\subsubsection{$\boldsymbol{Z^v}$-subproblem}

When $W^v,G,A,F$ are fixed, we can obtain the subproblem of Eq.(\ref{FGL-MSC}) with respect to $Z^v$:
\begin{equation} \label{Zp}
\begin{aligned}
    &\min_{Z^v} \sum_{v=1}^t \alpha||W^v-Z^v||_F^2 + \lambda \sum_{i=1}^n ||G_i^T - A_i^T \widetilde{Z}^i||^2_2\\
    &s.t.\quad Z^v \geq 0.
\end{aligned}
\end{equation}
For the sake of optimization, Eq.(\ref{Zp}) can be rewritten as:
\begin{equation} \label{Zpnew}
\begin{aligned}
    &\min_{\widetilde{Z}^i} \alpha||\widetilde{W}^i-\widetilde{Z}^i||_F^2 + \lambda||G_i^T - A_i^T \widetilde{Z}^i||^2_2\\
    &s.t.\quad \widetilde{Z}^i \geq 0,
\end{aligned}
\end{equation}
where $\widetilde{W}^i$ and $\widetilde{Z}^i$ are obtained in the same form as in Eq.(\ref{trans}).
Setting the first-order derivative of Eq.(\ref{Zpnew}) with respect to $\widetilde{Z}^i$ to zero, it yields:
\begin{equation} \label{solveZ}
    \widetilde{Z}^i=(\alpha I+\lambda A_iA_i^T)^{-1}(\alpha\widetilde{W}^i+\lambda A_iG_i^T).
\end{equation}
The sparse graphs $Z^v$ can be obtained by $trans^{-1}(\widetilde{Z})$. In order for the new $Z^v$ to behave as graphs, normalization and symmetrization are used.

\subsubsection{$\boldsymbol{G}$-subproblem}

When $W^v,Z^v,A,F$ are fixed, we can obtain the subproblem of Eq.(\ref{FGL-MSC}) with respect to $G$:
\begin{equation} \label{Gp}
\begin{aligned}
    &\min_{G} \lambda \sum_{i=1}^n ||G_i^T - A_i^T \widetilde{Z}^i||^2_2 +\gamma ||G||_F^2 + \eta Tr(F^TLF)\\
    &s.t.\quad G \geq 0, G\boldsymbol{1}=\boldsymbol{1}.
\end{aligned}
\end{equation}

To solve Eq.(\ref{Gp}), we use an equality:
\begin{equation} \label{tr}
    Tr(F^TLF)=\sum_{i,j=1}^n \frac{1}{2}||F_i-F_j||_2^2 g_{ij}.
\end{equation}
We denote $H$ as the distance matrix of F, that is, $h_{ij} = ||F_i-F_j||_2^2$. Eq.(\ref{Gp}) can be reformulated as:
\begin{equation} \label{Gip}
\begin{aligned}
    &\min_{G_i} (\lambda+\gamma) G_iG_i^T +2(-\lambda A_i^T \widetilde{Z}^i+\frac{\eta}{4}H_i^T)G_i\\
    &s.t.\quad G_i \geq 0, G_i\boldsymbol{1}=1.
\end{aligned}
\end{equation}

The problem (\ref{Gip}) can be solved as the Euclidean projection problem:
\begin{equation} \label{Eu}
\begin{aligned}
    &\min_{G_i} ||G_i + (\frac{-4\lambda A_i^T \widetilde{Z}^i+\eta H_i^T}{4(\lambda+\gamma)})||_2^2\\
    &s.t.\quad G_i \geq 0, G_i\boldsymbol{1}=1.
\end{aligned}
\end{equation}
The augmented Lagrangian function of Eq.(\ref{Eu}) is:
\begin{equation} \label{Lg}
\begin{aligned}
    &L(G_i,\sigma_1,\sigma_2)=||G_i + (\frac{-4\lambda A_i^T \widetilde{Z}^i+\eta H_i^T}{4(\lambda+\gamma)})||_2^2\\
    &\qquad + \sigma_1(G_i\boldsymbol{1}-1)+\sigma_2 G_i,
\end{aligned}
\end{equation}
where $\sigma_{1,2} \textgreater 0$ are the Lagrangian multipliers.

Generally, assume that the optimal $G_i$ has only $m$ non-negative elements. Let $Q_i^T=(\eta H_i^T-4\lambda A_i^T \widetilde{Z}^i)$, and be ordered from small to large, we can approximately set the $G$ as follows:
\begin{equation} \label{solveG}
    G_{ij}= \frac{Q_{i,m+1}-Q_{i,j}}{mQ_{i,m+1}-\sum_{j=1}^m Q_{i,j}}.
\end{equation}

Using Eq.(\ref{solveG}), we replace the harder-to-tune parameter $\gamma$ in the original problem with an integer $m$. Meanwhile, the average degree of a learned graph $G$ is controllable, i.e., only $m$ neighbors for each node are reported in $G$.
\subsubsection{$\boldsymbol{A}$-subproblem}

When $W^v,Z^v,G,F$ are fixed, we can obtain the subproblem of Eq.(\ref{FGL-MSC}) with respect to $A$:
\begin{equation} \label{Ap}
\begin{aligned}
    &\min ||G_i^T - A_i^T \widetilde{Z}^i||^2_2\\
    &s.t.\quad A_i \geq 0, A_i^T \boldsymbol{1}=1.
\end{aligned}
\end{equation}
After reorganization, Eq.(\ref{Ap}) can be written as:
\begin{equation} \label{Apnew}
\begin{aligned}
    &\min ||A_i^T T^i||^2_2\\
    &s.t.\quad A_i \geq 0, A_i^T \boldsymbol{1}=1,
\end{aligned}
\end{equation}
where $T^i=(\boldsymbol{1}G_i^T - \widetilde{Z}^i)\in R^{t \times n}$. Setting the first-order derivative of Eq.(\ref{Apnew}), it yields:
\begin{equation} \label{solveA}
    A_i= \frac{(T^i(T^i)^T)^{-1}\boldsymbol{1}}{\boldsymbol{1}^T (T^i(T^i)^T)^{-1}\boldsymbol{1}}.
\end{equation}

\subsubsection{$\boldsymbol{F}$-subproblem}

Given $W^v,Z^v,G,A$, the subproblem of $F$ becomes problem (\ref{spectral}). The optimal $F$ is formed by the $c$ eigenvectors of $L$ corresponding to the $c$ smallest eigenvalues.

\subsection{Discussions}

\begin{algorithm}[tb]
\caption{The general algorithm of FGL-MSC}
\label{alg:algorithm}
\raggedright\textbf{Input}: $t$ views data $\{X^1,\cdots,X^t\}$ with $X^v \in R^{d_v \times n}$, cluster number $c$.\\
\raggedright\textbf{Parameter}: $\alpha, \lambda, \eta \textgreater 0$, and the number of neighbors in $G$ is set to 10.\\
\raggedright\textbf{Output}: the unified graph $G$, the fine-grained fusion weight matrix $A$, the clustering representation $F$.
\begin{algorithmic}[1] 
\STATE Initialize the weight $a_{ij}=\frac{1}{t}$. 
\STATE Compute $Z^v$ by solving Eq.(\ref{CAN})., and $G$ is the average of all $Z^v$. 
\STATE Compute $F$ by solving Eq.(\ref{spectral}).
\WHILE{convergence condition does not meet}
\STATE Update $W^v$ by Eq.(\ref{solveW}).
\STATE Update $Z^v$ by Eq.(\ref{solveZ}).
\STATE Update $G$ by Eq.(\ref{solveG}).
\STATE Update $A$ by Eq.(\ref{solveA}).
\STATE Update $F$ by solving Eq.(\ref{spectral}).
\ENDWHILE
\STATE Apply $K$-means to $F$.
\end{algorithmic}
\end{algorithm}

Here we can give a clear interpretation of the mechanism of the proposed graph learning method. The initialization of graph $G$ in Eq.(\ref{FGL-MSC}) is defined as a mean of all the adaptive neighbors graphs generated from each view. During the optimization, the refined graph $\widetilde{Z}^i$ is learned to strike a balance between the informative graph $\widetilde{W}^i$ and the unified graph $G$ by Eq.(\ref{solveZ}). Next, a change in $Z$ leads to a change in the corresponding position of $G$ by Eq.(\ref{solveG}). According to Eq.(\ref{solveA}), the higher the degree of match with $G$, the higher weights the view will gain, whereas the effects resulting from redundant structures will be gradually digested and diluted. Finally, the unified graph $G_{ij}$ is given in the form of probabilities by Eq.(\ref{solveG}), the value of which depends on two aspects, the weighted sum of the edges between $i$ and $j$ in all refined graphs, and the Euclidean distance of their clustering representation through spectral decomposition. In addition, the optimization algorithm we use can also control the sparsity of the unified graph by adjusting the parameter $m$. Here we theoretically analyze the convergence and computational complexity of the optimization algorithm, $i.e.$, Algorithm \ref{alg:algorithm}.

\subsubsection{Convergence Analysis}

The convergence of Algorithm \ref{alg:algorithm} is given by Theorem 1.

\emph{Theorem 1:} The alternate updating strategy in Algorithm \ref{alg:algorithm} monotonically decrease the value of Eq.(\ref{FGL-MSC}) in each iteration until convergence.

\emph{Proof:} Firstly, the convergence of Eq.(\ref{solveW}) is provided in \cite{LEnewS}. Let $l$ denotes the iterating times, the overall objective function $\mathcal{L}(\mathcal{W}, \mathcal{Z}, G, A, F)$ monotonically decreases:
\begin{equation} \label{converge1}
    \mathcal{L}((\mathcal{W})^{l+1}, \mathcal{Z}, G, A, F) \leq \mathcal{L}((\mathcal{W})^{l}, \mathcal{Z}, G, A, F).
\end{equation}

Next, The Hessian matrix of Eq.(\ref{Zpnew}) is:
\begin{equation} \label{converge2}
    \frac{\partial^2 \mathcal{L}(\widetilde{Z}^i)}{\partial (\widetilde{Z}^i)^2} = \alpha I + \lambda A_i A_i^T.
\end{equation}

It is clear that $\alpha||\widetilde{W}^i-\widetilde{Z}^i||_F^2 + \lambda||G_i^T - A_i^T \widetilde{Z}^i||^2_2 \geq 0$ and Eq.(\ref{converge2}) is positive semi-definite. Therefore Eq.(\ref{Zpnew}) is a convex function. The overall objective function monotonically decreases:
\begin{equation} \label{converge3}
    \mathcal{L}(\mathcal{W}, (\mathcal{Z})^{l+1}, G, A, F) \leq \mathcal{L}(\mathcal{W}, (\mathcal{Z})^{l}, G, A, F).
\end{equation}

When others are fixed and updating $G$ or $A$, Eq.(\ref{Eu}) and Eq.(\ref{Ap}) are both convex functions because the second-order derivatives of them with respect to each variant are greater than 0. So the overall objective function monotonically decreases:
\begin{equation} \label{converge4}
\begin{aligned}
    &\mathcal{L}(\mathcal{W}, \mathcal{Z}, G^{l+1}, A, F) \leq \mathcal{L}(\mathcal{W}, \mathcal{Z}, G^{l}, A, F)\\
    &\mathcal{L}(\mathcal{W}, \mathcal{Z}, G, A^{l+1}, F) \leq \mathcal{L}(\mathcal{W}, \mathcal{Z}, G, A^{l}, F).
\end{aligned}
\end{equation}

Finally, according to \cite{eigenvalue}, the Hessian matrix of the Lagrangian function of Eq.(\ref{spectral}) is also positive semi-definite. So Eq.(\ref{spectral}) is a convex function, and the overall objective function monotonically decreases:
\begin{equation} \label{converge4}
    \mathcal{L}(\mathcal{W}, \mathcal{Z}, G, A, F^{l+1}) \leq \mathcal{L}(\mathcal{W}, \mathcal{Z}, G, A, F^{l}).
\end{equation}

This completes the proof. 

\subsubsection{Computational Complexity Analysis}

The first step of FGL-MSC is to solve Eq.(\ref{CAN}), whose cost is $O(n^2)$. The second step is to solve Eq.(\ref{solveW}), which costs $O(n^3)$ based on \cite{LEnewS}. In the third and the fourth step of FGL-MSC, we need $O((t^2+tn+t^2n)n)$ and $O(n^2+tn^2)$ to calculate $\mathcal{Z}$ and $G$, respectively. The fifth step is to solve Eq.(\ref{solveA}), we need $O(tn)$ to calculate $T^i$, and the cost of Eq.(\ref{solveA}) is $O(t^2n+t^2)$. At last, the eigendecomposition problem \ref{spectral} cost $O(cn^2)$. Therefore, the total computational complexity of FGL-MSC is as follows:
\begin{equation} \label{converge4}
    O(n^2 + l(n^3 + (1+3t+2t^2+c)n^2 + 2t^2 n)),
\end{equation}
where $l$ is the iteration of the above updating steps. Because of $n \gg t$, $n \gg c$, and $n \gg l$, the main computational complexity of FGL-MSC is at the same level as most multi-view clustering methods.
The computational complexity mainly locates in computing (\ref{solveW}). Due to the existence of Lasso regularization, more efficient optimization strategies should be investigated.

\section{Experiment}

In this section, we carry out clustering experiments on 8 benchmark datasets, comparing FGL-MSC with 10 related multi-view clustering methods in terms of three performance evaluation metrics. All the experiments are performed using MATLAB 2021b on a Linux Server with an Intel Xeon 2.10GHz CPU and 128GB RAM.

\subsection{Datasets and Experimental Setup}

To demonstrate the efficacy of the proposed framework FGL-MSC totally, we conduct experiments on five small-scale(the number of nodes is less than 2000) benchmark datasets and three large-scale datasets.
We normalized the data in the range $[0,1]$, taking into account the metric differences between different forms of features \cite{CAI}. The detailed information of small-scale datasets is as follows:

\textbf{MSRC-v1} \cite{MSRC} dataset contains 210 nodes with 7 classes: tree, airplane, face, car, building, bicycle, and cow. Each class has 30 nodes. The following 4 views are available: CM, GISTM, LBP, and CENT.

\textbf{BBCSport}\footnote{\url{http://mlg.ucd.ie/datasets/}} dataset is composed of news articles in 5 topical areas from the BBC website, which is associated with 2 views.

\textbf{WebKB}\footnote{\url{http://vikas.sindhwani.org/MR.zip}} dataset is a web page classification data set. It contains 203 nodes with binary classes. The following 3 views are available: The Web page text content called PAGE, the anchor text on links called LINK, and the 'Page-Link' content.

\textbf{100leaves}\footnote{\url{https://archive.ics.uci.edu/ml/datasets/Onehundred+plant+species+leaves+data+set}} dataset contains 1600 samples with 100 plant species and each class has 16 samples. There are 3 published features can be used for clustering: shape descriptor, texture histogram, and fine-scale margin.

\textbf{ORL}\footnote{\url{http://www.uk.research.att.com/facedatabase.html}} dataset contains 400 face images with 40 classes. Three views can be used: 6750 dimension Gabor, 4096 dimension intensity, and 3304 dimensions LBP.

Likewise, Three large-scale datasets are as follows:

\textbf{Caltech101-20}\footnote{\url{http://www.vision.caltech.edu/ImageDatasets/Caltech101/}} (Caltech20) dataset contains 2386 nodes with 20 classes. 6 views can be used: Gabor, Wavelet moments, Cenhist, HOG, GIST, and LBP.

\textbf{Scene-15} (scene15) \cite{scene15} dataset consists of 4485 nodes that cover 15 categories: suburb, bedroom, kitchen, industrial, living, room, forest, inside city, coast, office, highway, tall building, mountain, street, open country, and store. 3 views are available.

\textbf{Hdigit}\footnote{\url{https://cs.nyu.edu/roweis/data.html}} dataset contains 10000 nodes. This handwritten digits (0-9) data set is from two sources, i.e., MNIST Handwritten Digits and USPS Handwritten Digits.

\begin{table*}
\centering
\caption{Clustering results on small-scale datasets (mean ± standard deviation)}
\begin{tabular}{llccccc}
\toprule
Methods & Metrics & MSRC-v1 & BBCSport & WebKB & ORL & 100leaves \\
\midrule
SC(best) & ACC     &  0.6914$\pm$0.01  &  0.3675$\pm$0.00  &  0.6207$\pm$0.00  &  0.7388$\pm$0.03  &  0.6261$\pm$0.02   \\
         & NMI     &  0.6861$\pm$0.03  &  0.0368$\pm$0.01  &  0.2100$\pm$0.00  &  0.9079$\pm$0.01  &  0.8232$\pm$0.01  \\
		& ARI     &  0.5589$\pm$0.04  &  0.0046$\pm$0.00  &  0.1416$\pm$0.00  &  0.6383$\pm$0.06  &  0.4517$\pm$0.04  \\
\midrule
AMGL     & ACC     &  0.6991$\pm$0.06  &  0.3585$\pm$0.00  &  0.5704$\pm$0.01  &  0.7293$\pm$0.02  &  0.7706$\pm$0.02  \\
         & NMI     &  0.6594$\pm$0.03  &  0.0259$\pm$0.00  &  0.1322$\pm$0.03  &  0.8914$\pm$0.01  &  0.9182$\pm$0.01  \\
		& ARI     &  0.5107$\pm$0.06  &  0.0008$\pm$0.00  &  0.0726$\pm$0.02  &  0.5479$\pm$0.05  &  0.6176$\pm$0.07  \\
\midrule
MLAN     & ACC     &  0.7381$\pm$0.00  &  0.7962$\pm$0.00  &  0.7251$\pm$0.00  &  0.685$\pm$0.00  &  0.8731$\pm$0.00  \\
         & NMI     &  0.7515$\pm$0.00  &  \textbf{0.7304$\pm$0.00}  &  0.4022$\pm$0.00  &  0.8312$\pm$0.00  &  0.9498$\pm$0.00  \\
	     & ARI     &  0.6441$\pm$0.00  &  0.6554$\pm$0.00  &  0.3732$\pm$0.00  &  0.3316$\pm$0.00  &  0.8242$\pm$0.00  \\
\midrule
MVGL      & ACC     &  0.7314$\pm$0.00  &  0.7946$\pm$0.00  &  0.6882$\pm$0.00  &  0.8073$\pm$0.00  &  0.8562$\pm$0.00   \\
          & NMI     &  0.7025$\pm$0.00  &  0.7012$\pm$0.00  &  0.3984$\pm$0.00  &  0.8169$\pm$0.00  &  0.9495$\pm$0.00  \\
		 & ARI     &  0.5973$\pm$0.00  &  0.6889$\pm$0.00  &  0.4724$\pm$0.00  &  0.6590$\pm$0.00  &  0.8637$\pm$0.00  \\
\midrule
GMC      & ACC     &  0.7476$\pm$0.00  &  0.8036$\pm$0.00  &  0.7685$\pm$0.00  &  0.6325$\pm$0.00  &  0.8238$\pm$0.00  \\
         & NMI     &  0.7143$\pm$0.00  &  0.7600$\pm$0.00    &  0.4387$\pm$0.00  &  0.8571$\pm$0.00  &  0.9292$\pm$0.00  \\
		& ARI     &  0.6161$\pm$0.00  &  0.6938$\pm$0.00  &  0.4676$\pm$0.00  &  0.3367$\pm$0.00  &  0.4974$\pm$0.00  \\
\midrule
LMSC     & ACC     &  0.6210$\pm$0.03  &  0.7925$\pm$0.07           &  0.5739$\pm$0.05           &  0.8133$\pm$0.03  &  0.7275$\pm$0.02  \\
         & NMI     &  0.5262$\pm$0.02  &  0.7144$\pm$0.03           &  0.2091$\pm$0.01           &  0.9224$\pm$0.01  &  0.8734$\pm$0.01  \\
		& ARI     &  0.4182$\pm$0.02  &  0.6999$\pm$0.07  &  0.2223$\pm$0.02  &  0.7590$\pm$0.03  &  0.6393$\pm$0.02  \\
\midrule
GFSC     & ACC     &  0.7405$\pm$0.04  &  0.3585$\pm$0.00           &  0.7966$\pm$0.04     &  0.6708$\pm$0.03  &  0.5795$\pm$0.06  \\
         & NMI     &  0.6674$\pm$0.02  &  0.0149$\pm$0.00           &  \textbf{0.5419$\pm$0.00}           &  0.8556$\pm$0.01  &  0.7264$\pm$0.07  \\
	    & ARI     &  0.5714$\pm$0.04  &  0.0007$\pm$0.00  &  0.4087$\pm$0.02  &  0.5436$\pm$0.04  &  0.4067$\pm$0.03  \\
\midrule
CGL      & ACC     &  0.7381$\pm$0.00  &  0.8006$\pm$0.05           &  0.5788$\pm$0.04           &  0.8348$\pm$0.02  &  0.9346$\pm$0.01  \\
         & NMI     &  0.6626$\pm$0.00  &  0.7286$\pm$0.02           &  0.2815$\pm$0.03           &  0.9189$\pm$0.01  &  0.9735$\pm$0.00  \\
		& ARI     &  0.5688$\pm$0.00  &  0.6951$\pm$0.04  &  0.2958$\pm$0.03  &  0.7695$\pm$0.02  &  0.9116$\pm$0.02  \\
\midrule
CNESE      & ACC     &  0.7481$\pm$0.00  &  0.7514$\pm$0.00           &  0.7608$\pm$0.00           &  0.7175$\pm$0.00  &  0.8756$\pm$0.00  \\
         & NMI     &  0.6916$\pm$0.02  &  0.6783$\pm$0.00           &  0.4195$\pm$0.00           &  0.8354$\pm$0.00  &  0.9353$\pm$0.00  \\
		& ARI     &  0.5962$\pm$0.00  &  0.6294$\pm$0.00  &  0.5259$\pm$0.00  &  0.5949$\pm$0.01  &  0.8204$\pm$0.01  \\
\midrule
CiMSC      & ACC     &  \textbf{0.7830$\pm$0.03}  &  0.7205$\pm$0.10           &  \textbf{0.8129$\pm$0.04}           &  0.8337$\pm$0.18  &  0.9547$\pm$0.11  \\
         & NMI     &  \textbf{0.7609$\pm$0.02}  &  0.6839$\pm$0.08           &  0.5264$\pm$0.04           &  0.8765$\pm$0.22  &  0.9646$\pm$0.07  \\
		& ARI     &  \textbf{0.7051$\pm$0.05}  &  0.6004$\pm$0.01  &  0.5230$\pm$0.01  &  0.7659$\pm$0.25  &  0.9163$\pm$0.06  \\
\midrule
FGL-MSC     & ACC     &  0.7619$\pm$0.00  &  \textbf{0.8042$\pm$0.05}  &  0.7256$\pm$0.04  &  \textbf{0.8463$\pm$0.00}  &  \textbf{0.9631$\pm$0.00}  \\
         & NMI     &  0.7144$\pm$0.00  &  0.6865$\pm$0.04  &  0.4021$\pm$0.02  &  \textbf{0.9286$\pm$0.00}  &  \textbf{0.9743$\pm$0.00}  \\
		& ARI      &  0.6074$\pm$0.00  &  \textbf{0.7013$\pm$0.09}  &  \textbf{0.5479$\pm$0.01}  &  \textbf{0.7771$\pm$0.00}  &  \textbf{0.9242$\pm$0.00}  \\
\bottomrule
\end{tabular}
\label{new_re}
\end{table*}

\begin{table}
\centering
\caption{Clustering results (ACC) on large-scale datasets (mean ± standard deviation)}
\begin{tabular}{lccc}
\toprule
Methods & Caltech20 & scene15 & Hdigit \\
\midrule
SC(best) &  0.3204$\pm$0.02  &  0.2856$\pm$0.02           &  0.6492$\pm$0.04  \\
\midrule
AMGL     &  0.5641$\pm$0.02  &  0.3145$\pm$0.02           &  0.9603$\pm$0.06  \\
\midrule
MLAN    &  0.5201$\pm$0.00  &  0.1523$\pm$0.00           &  0.8461$\pm$0.00  \\
\midrule
MVGL     &  0.5706$\pm$0.00  &  0.3675$\pm$0.00           &  0.9420$\pm$0.00  \\
\midrule
GMC      &  0.4564$\pm$0.00  &  0.1400$\pm$0.00           &  0.9681$\pm$0.00  \\
\midrule
LMSC      &  0.4643$\pm$0.04  &  0.3872$\pm$0.01           &  0.9642$\pm$0.00  \\
\midrule
GFSC     &  0.5776$\pm$0.03  &  0.3609$\pm$0.01           &  0.9762$\pm$0.00  \\
\midrule
CGL      &  0.5500$\pm$0.03  &  0.4186$\pm$0.01           &  0.9769$\pm$0.02  \\
\midrule
CNESE    &  0.4956$\pm$0.00  &  \textbf{0.4378$\pm$0.01}  &  0.9757$\pm$0.00  \\
\midrule
CiMSC    &  \textbf{0.5987$\pm$0.20}  &  0.3384$\pm$0.04  &  0.9707$\pm$0.07  \\
\midrule
FGL-MSC     &  0.5845$\pm$0.01  &  0.3734$\pm$0.03  &  \textbf{0.9782$\pm$0.00}  \\
\bottomrule
\end{tabular}
\label{large_results_acc}
\end{table}

\begin{table}
\centering
\caption{Clustering results (NMI) on large-scale datasets (mean ± standard deviation)}
\begin{tabular}{lccc}
\toprule
Methods & Caltech20 & scene15 & Hdigit \\
\midrule
SC(best) &  0.3961$\pm$0.01  &  0.2970$\pm$0.01           &  0.7242$\pm$0.02  \\
\midrule
AMGL    &  0.6013$\pm$0.04  &  0.3383$\pm$0.01           &  0.9752$\pm$0.02   \\
\midrule
MLAN     &  0.5385$\pm$0.00  &  0.1538$\pm$0.00           &  0.9158$\pm$0.00  \\
\midrule
MVGL   &  0.5317$\pm$0.00  &  0.3178$\pm$0.00           &  0.8905$\pm$0.00  \\
\midrule
GMC     &  0.4809$\pm$0.00  &  0.1016$\pm$0.00           &  \textbf{0.9939$\pm$0.00}  \\
\midrule
LMSC    &  0.6041$\pm$0.02  &  0.3671$\pm$0.01           &  0.9732$\pm$0.00  \\
\midrule
GFSC    &  0.5124$\pm$0.06  &  0.3304$\pm$0.02           &  0.9500$\pm$0.00  \\
\midrule
CGL     &  \textbf{0.6540$\pm$0.02}  &  0.3980$\pm$0.01           &  0.9496$\pm$0.03  \\
\midrule
CNESE    &  0.5645$\pm$0.03  &  0.3209$\pm$0.01  &  0.9300$\pm$0.01  \\
\midrule
CiMSC    &  0.6326$\pm$0.10  &  \textbf{0.4167$\pm$0.07}  &  0.9283$\pm$0.02  \\
\midrule
FGL-MSC    &  0.5757$\pm$0.00  &  0.3329$\pm$0.01  &  0.9418$\pm$0.00  \\
\bottomrule
\end{tabular}
\label{large_results_nmi}
\end{table}

\begin{table}
\centering
\caption{Clustering results (ARI) on large-scale datasets (mean ± standard deviation)}
\begin{tabular}{lccc}
\toprule
Methods & Caltech20 & scene15 & Hdigit \\
\midrule
SC(best) &  0.1670$\pm$0.02   &  0.1217$\pm$0.01   &  0.5614$\pm$0.02  \\
\midrule
AMGL    &  0.3707$\pm$0.06  &  0.1358$\pm$0.01  &  0.9457$\pm$0.06  \\
\midrule
MLAN     &  0.1925$\pm$0.00 &  0.0054$\pm$0.00  &  0.8430$\pm$0.00  \\
\midrule
MVGL     &  \textbf{0.4596$\pm$0.00} &  0.1463$\pm$0.00  &  0.9420$\pm$0.00  \\
\midrule
GMC      &  0.1284$\pm$0.00  &  0.0042$\pm$0.00  &  0.9458$\pm$0.00  \\
\midrule
LMSC    &  0.3397$\pm$0.03  &  0.2138$\pm$0.01  &  0.9493$\pm$0.00  \\
\midrule
GFSC     &  0.4003$\pm$0.07  &  0.1549$\pm$0.01  &  0.9271$\pm$0.00  \\
\midrule
CGL    &  0.4408$\pm$0.03  &  \textbf{0.2391$\pm$0.01}  &  0.9507$\pm$0.02  \\
\midrule
CNESE    &  0.2564$\pm$0.01  &  0.2047$\pm$0.00  &  0.8800$\pm$0.04  \\
\midrule
CiMSC    &  0.4461$\pm$0.05  &  0.2165$\pm$0.01  &  0.9242$\pm$0.02  \\
\midrule
FGL-MSC  &  0.2641$\pm$0.01  &  0.1762$\pm$0.01  &  \textbf{0.9521}$\pm$0.00  \\
\bottomrule
\end{tabular}
\label{large_results_ari}
\end{table}

The proposed FGL-MSC is evaluated in three wide evaluation metrics, including accuracy (ACC), normalized mutual information (NMI), and adjusted Rand index (ARI). For all measures, the higher the scores, the better the model performs.

We compare FGL-MSC with 10 state-of-the-art multi-view clustering methods.
\textbf{Spectral Clustering (SC)} \cite{SC1} performs the standard spectral clustering in each view. The best results of them are reported. \textbf{AMGL} \cite{AMGL} proposes a parameter-free framework to implement the multi-view SC model. \textbf{MLAN} \cite{MLAN} extends the adaptive neighbors graph proposed by \cite{CAN} to multi-view graph learning. \textbf{MVGL} \cite{MVGL} proposes a novel fine-grained graph fusion pattern for a set of adaptive neighbors graphs generated from each view. \textbf{GMC} \cite{GMC} conducts a general framework that ensembles graph learning, graph fusion, and clustering. \textbf{LMSC} \cite{LMSC} is a representative multi-view subspace clustering method that seeks a common latent representation from the original features and employs subspace clustering. \textbf{GFSC} \cite{GFSC} combines subspace clustering and graph fusion to create a jointly multi-view clustering framework. \textbf{CGL} \cite{CGL} proposes a graph learning method to conduct clustering by simultaneously learning spectral embedding matrices and low-rank tensor representation. \textbf{CNESE} \cite{CNESE} is a constrained multi-view spectral clustering model via integrating non-negative embedding and spectral embedding. \textbf{CiMSC} \cite{CiMSC} is a consistency-induced multi-view subspace clustering framework. All source codes are downloaded from their author's pages, and their detailed configurations are referenced from the experimental setups provided in their papers. For a fair comparison, we run each experiment 10 times to report the mean and standard deviation.

\subsection{Clustering Results}

The experimental results (ACC, NMI, and ARI) of different methods on small-scale datasets are reported in Table \ref{new_re}, respectively. The results on large-scale datasets are presented in Table \ref{large_results_acc}, \ref{large_results_nmi}, and \ref{large_results_ari}, respectively. The best results are marked in \textbf{boldface}. According to these results, we have the following observations: 

\begin{enumerate}
	\item Comparing the basic SC with others, the best result of SC tends to lag behind multi-view methods. It illustrates the importance of investigating multi-view models, which exploit information that cannot be detected in a single view. 
    \item On the one hand, FGL-MSC shows competitive performance on the above datasets. In terms of ACC and ARI, our model outperforms all baseline models except CiMSC on MSRC-v1, BBCSport, ORL, 100leaves, and Hdigit. For WebKB, our FGL-MSC exceeds AMGL, MLAN, MVGL, LMSC, and CGL concerning all metrics. For Caltech20, our method performs better than MLAN, GMC, and CNESE concerning all metrics. For scene15, FGL-MSC outperforms AMGL, MLAN, MVGL, GMC, and GFSC concerning all metrics. On the other hand, fine-grained weights can cause the model to rely more on the quality of the multi-view data. But on large-scale datasets, our model still gets the best results in terms of ACC on Hdigit. It also exceeds the performances of AMGL, MLAN, GMC, and GFSC on most metrics. Note that despite better results of LMSC, CGL, and CiMSC, their time and space consumption is more substantial than our model. These results can prove that our strategy is successful.
    \item Our model performs better than MVGL for almost all measures on every dataset. The difference between FGL-MSC and MVGL is that our model builds a joint learning framework for the single graph on each view, the unified graph, the fusion weights, and the clustering representation, which enhance each other. The results show the benefits of our model.
    \item Our model is ahead of GFSC for most measures on six datasets. Though GFSC exploits a novel graph-level fusion method, the learned graph is probably not as accurate as the fine-grained one in most cases. This fully demonstrates the efficacy of our model.
\end{enumerate}

\subsection{Parameter Analysis}

\begin{figure*}
\centering
\vspace{0.1in}
\subfigtopskip=-1pt 
\subfigcapskip=-5pt 
\subfigure[$\alpha=0.01$]{
	\label{l_1} 
	\includegraphics[width=0.3\linewidth]{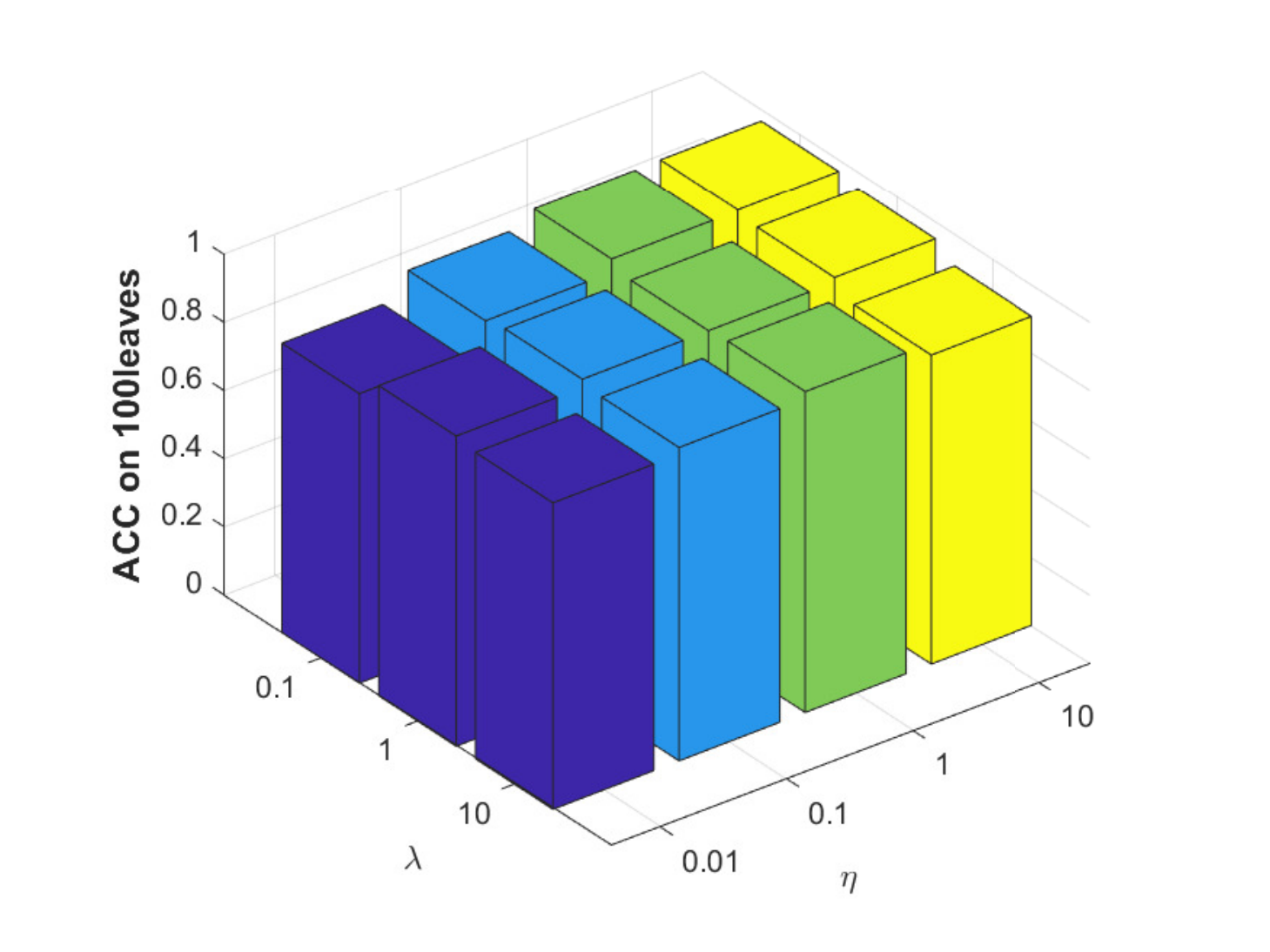}}
\hspace{0.1in}
\subfigure[$\alpha=0.1$]{
	\label{l_2}
	\includegraphics[width=0.3\linewidth]{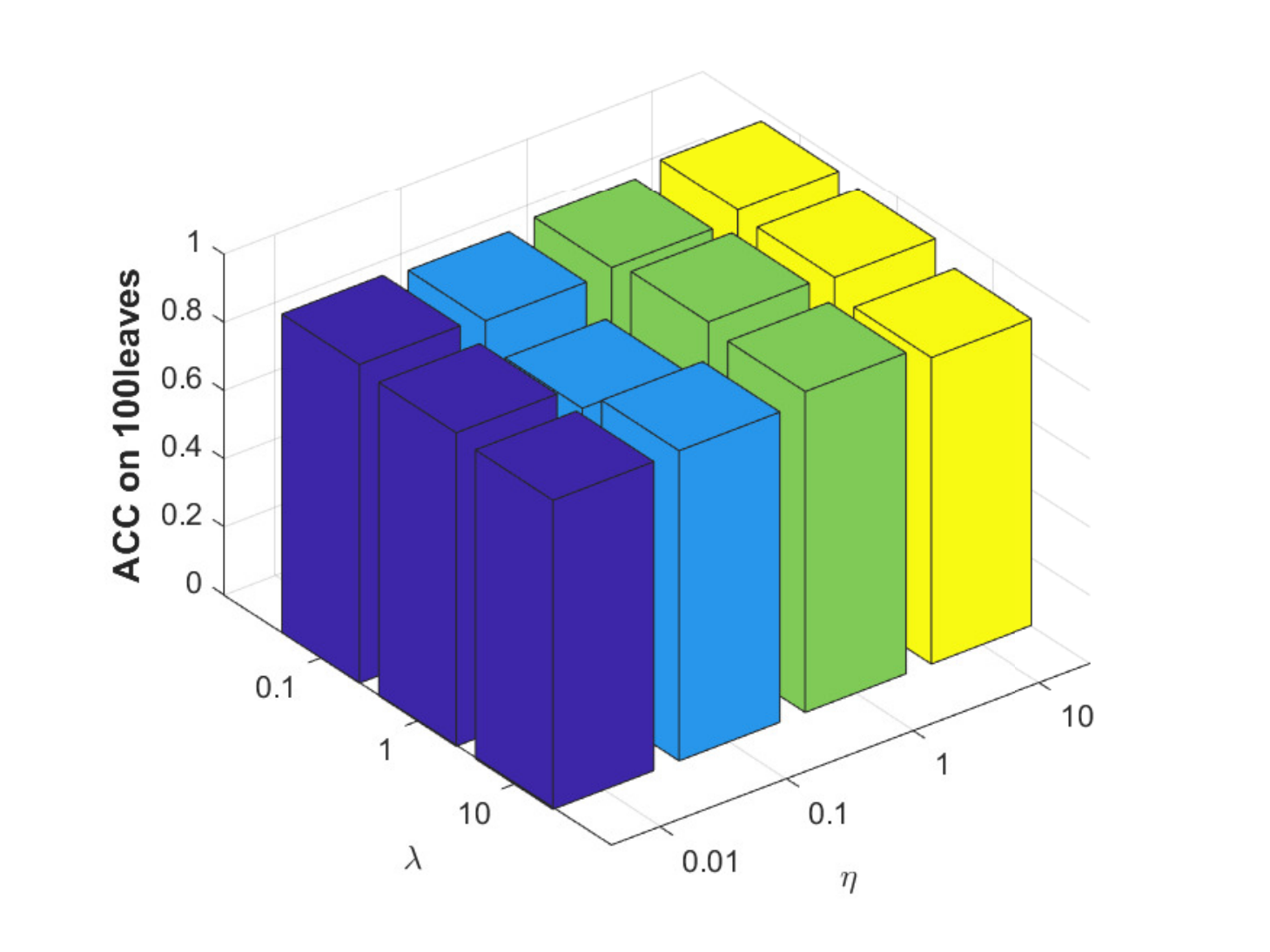}}
\hspace{0.1in}
\subfigure[$\alpha=1$]{
	\label{l_3}
	\includegraphics[width=0.3\linewidth]{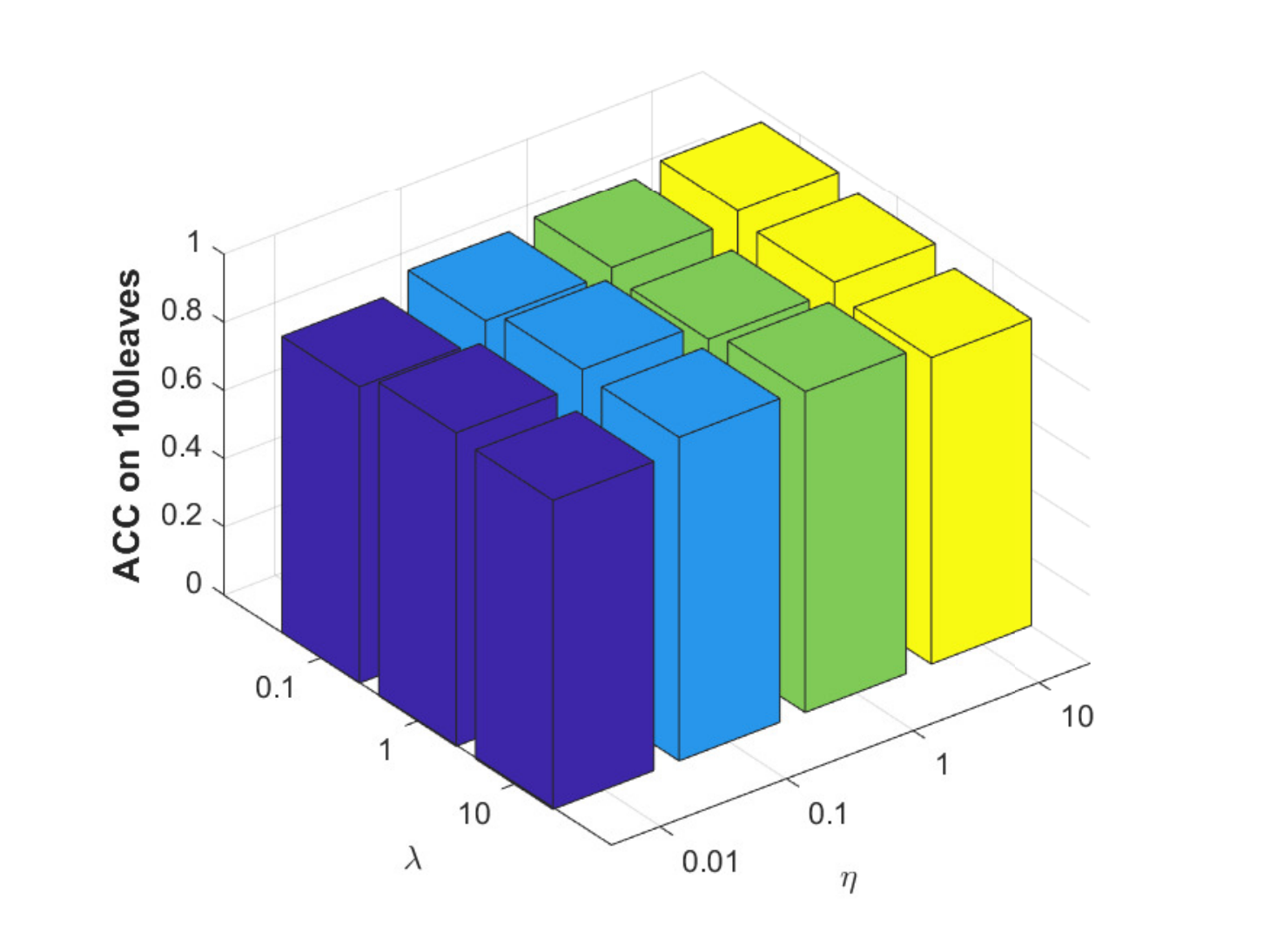}}

\subfigure[$\alpha=0.01$]{
	\label{m_1} 
	\includegraphics[width=0.3\linewidth]{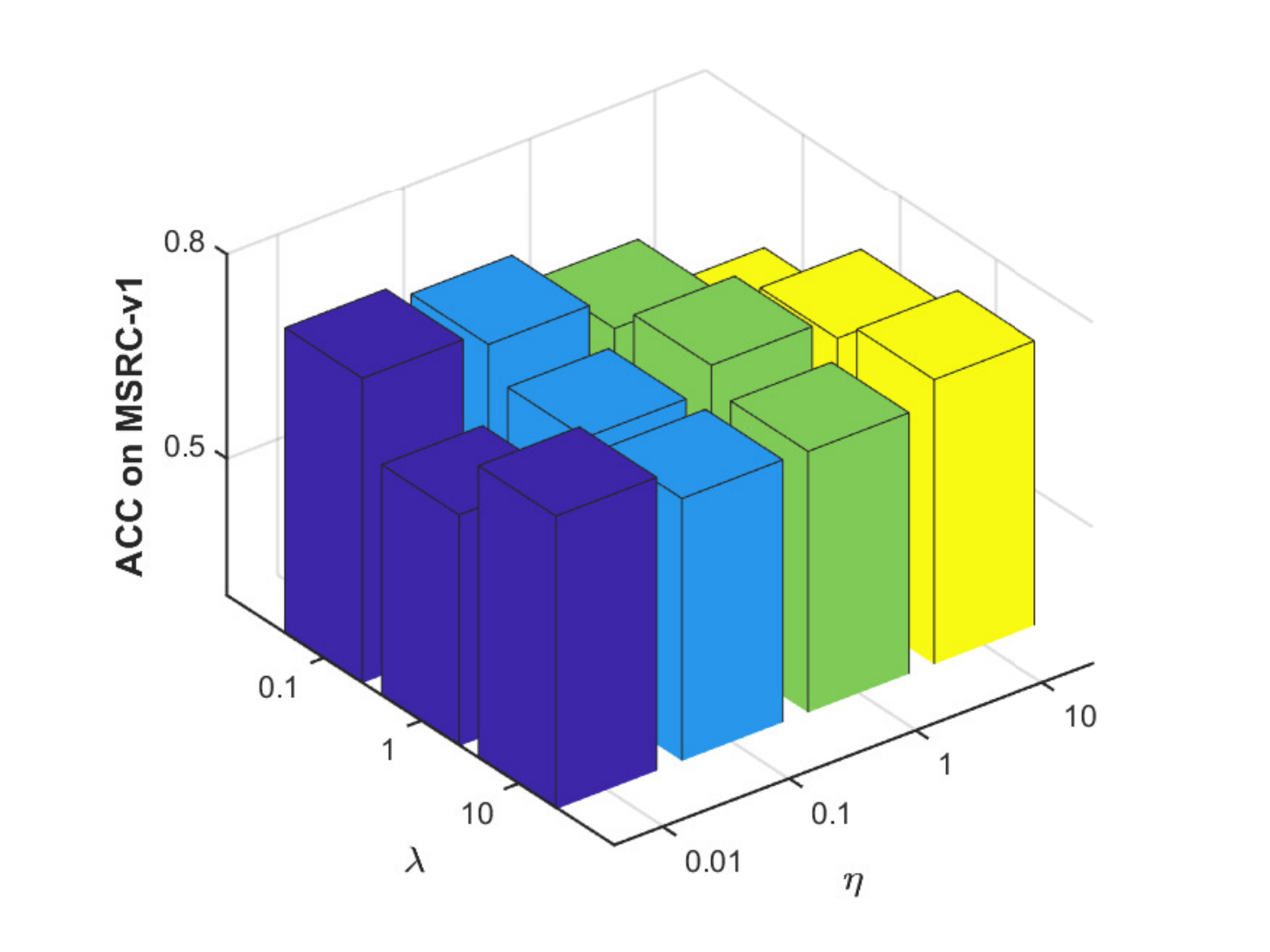}}
\hspace{0.1in}
\subfigure[$\alpha=0.1$]{
	\label{m_2}
	\includegraphics[width=0.3\linewidth]{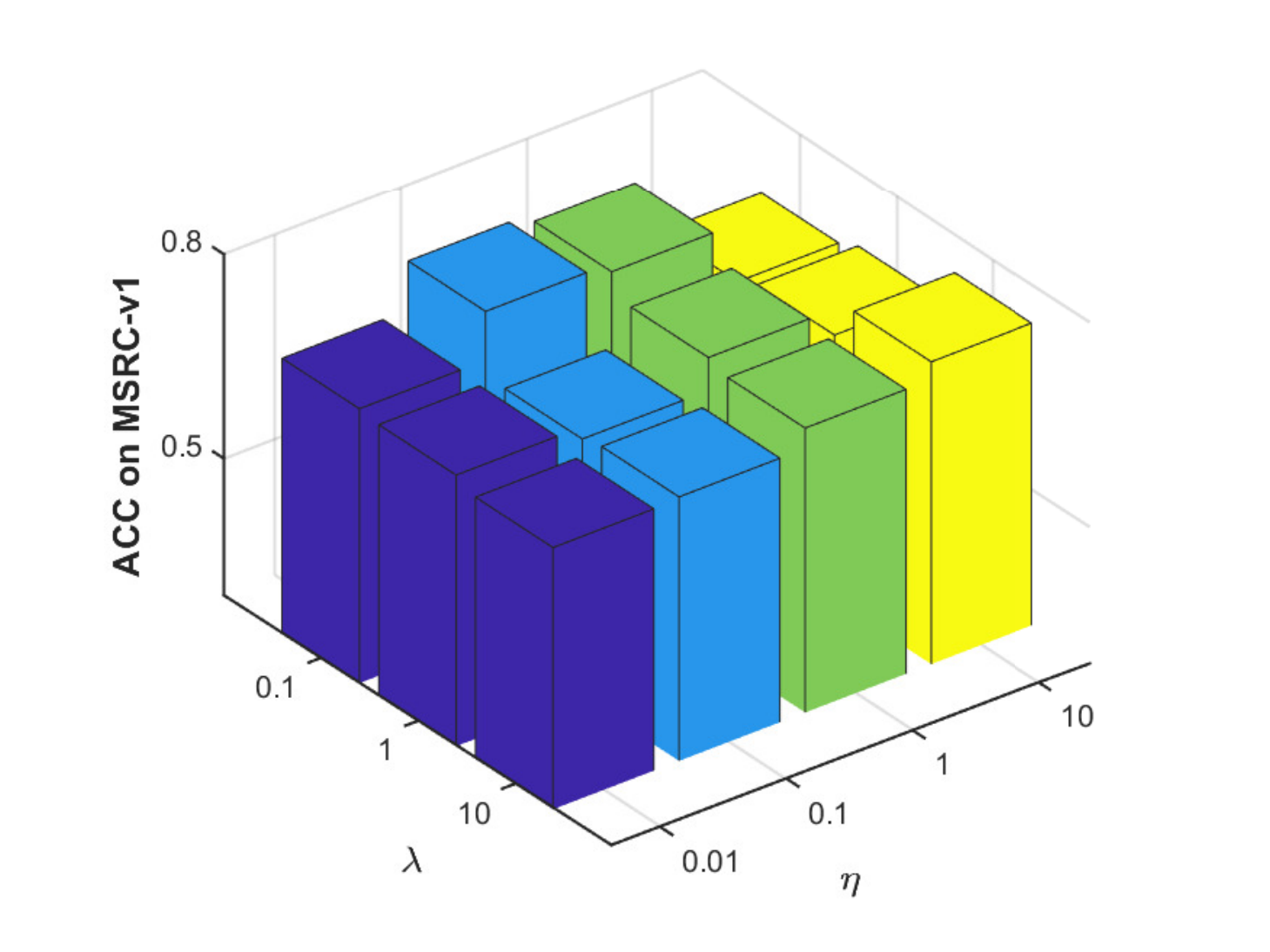}}
\hspace{0.1in}
\subfigure[$\alpha=1$]{
	\label{m_3}
	\includegraphics[width=0.3\linewidth]{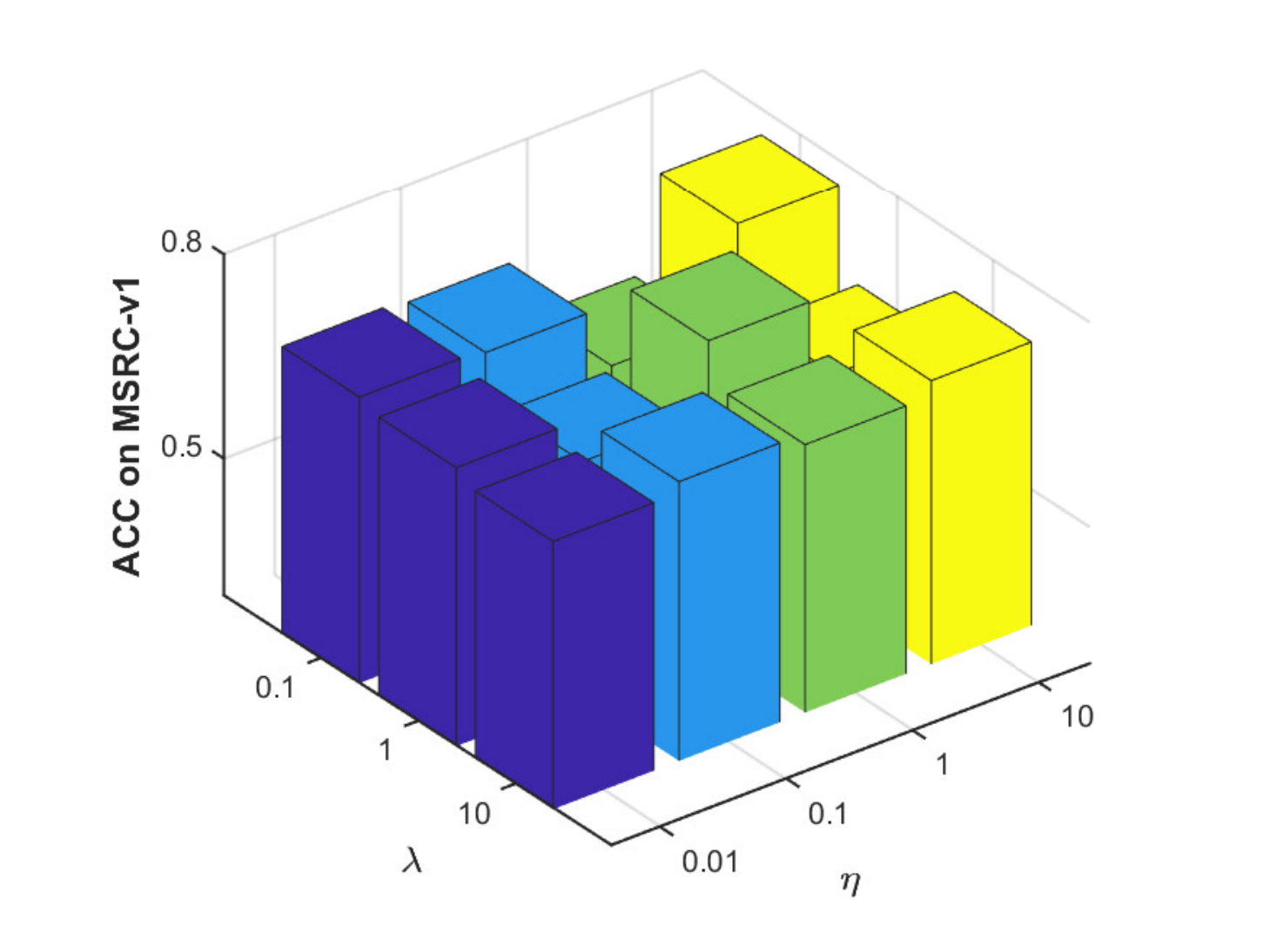}}

\subfigure[$\alpha=0.01$]{
	\label{o_1} 
	\includegraphics[width=0.3\linewidth]{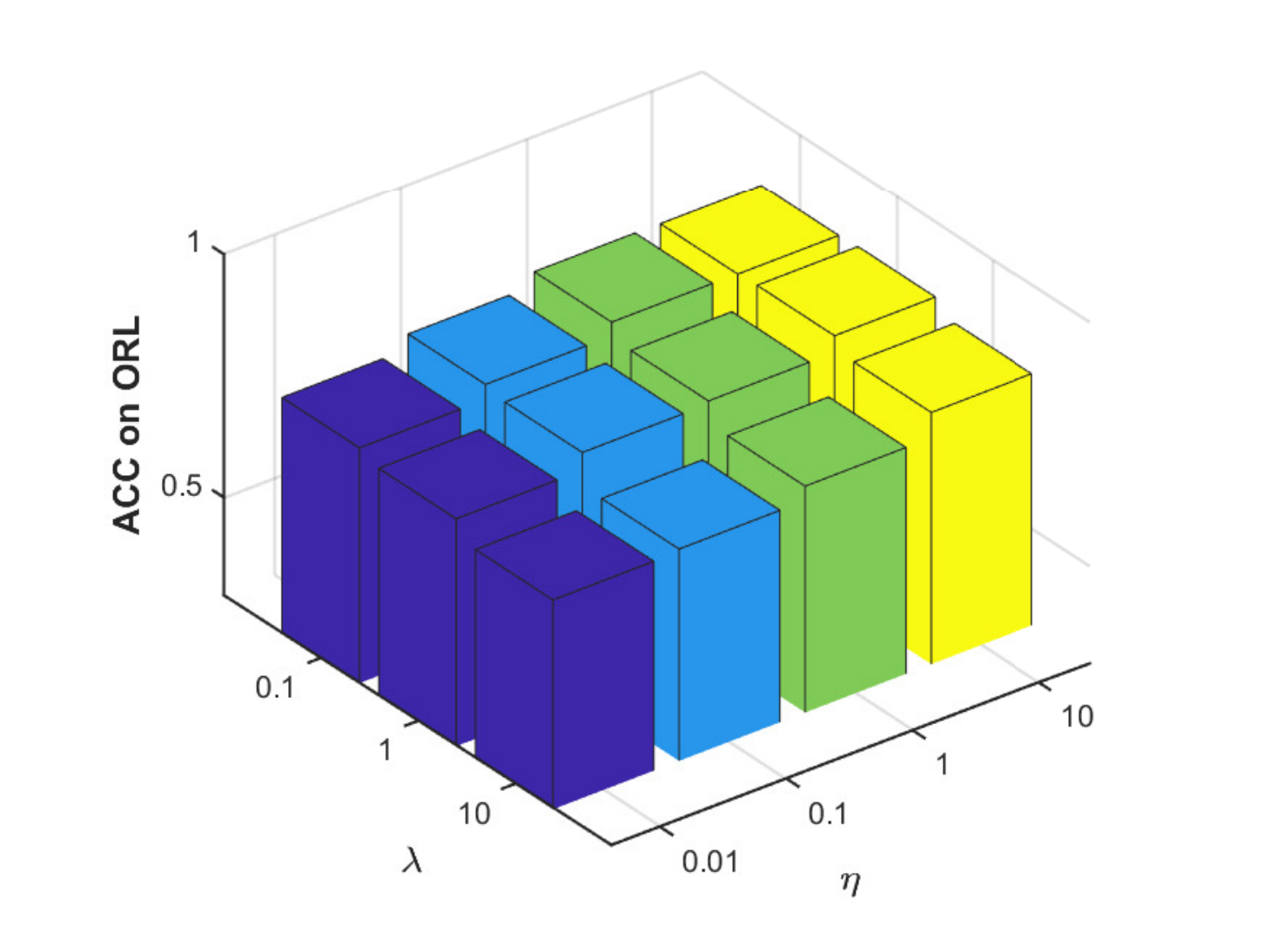}}
\hspace{0.1in}
\subfigure[$\alpha=0.1$]{
	\label{o_2}
	\includegraphics[width=0.3\linewidth]{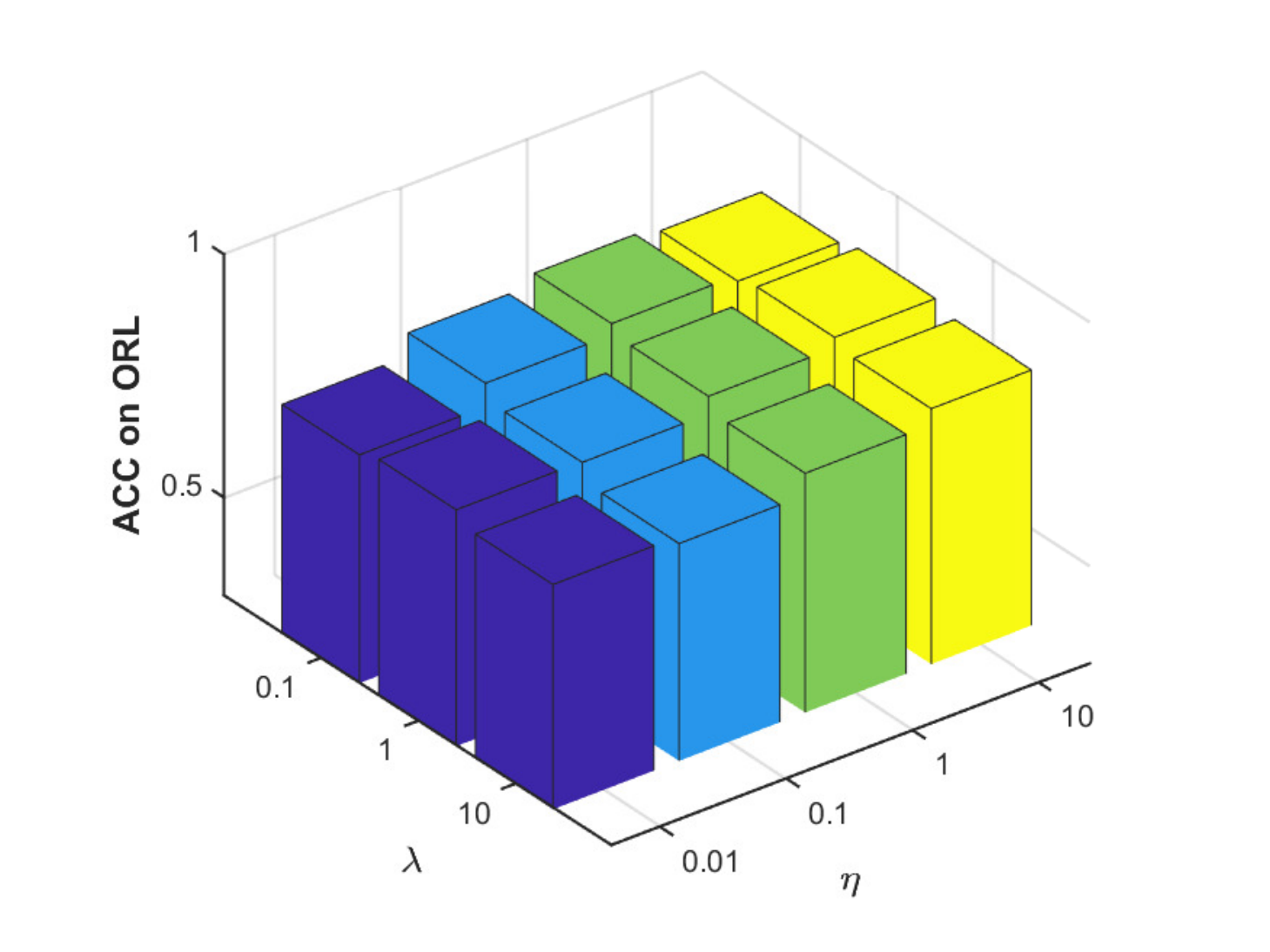}}
\hspace{0.1in}
\subfigure[$\alpha=1$]{
	\label{o_3}
	\includegraphics[width=0.3\linewidth]{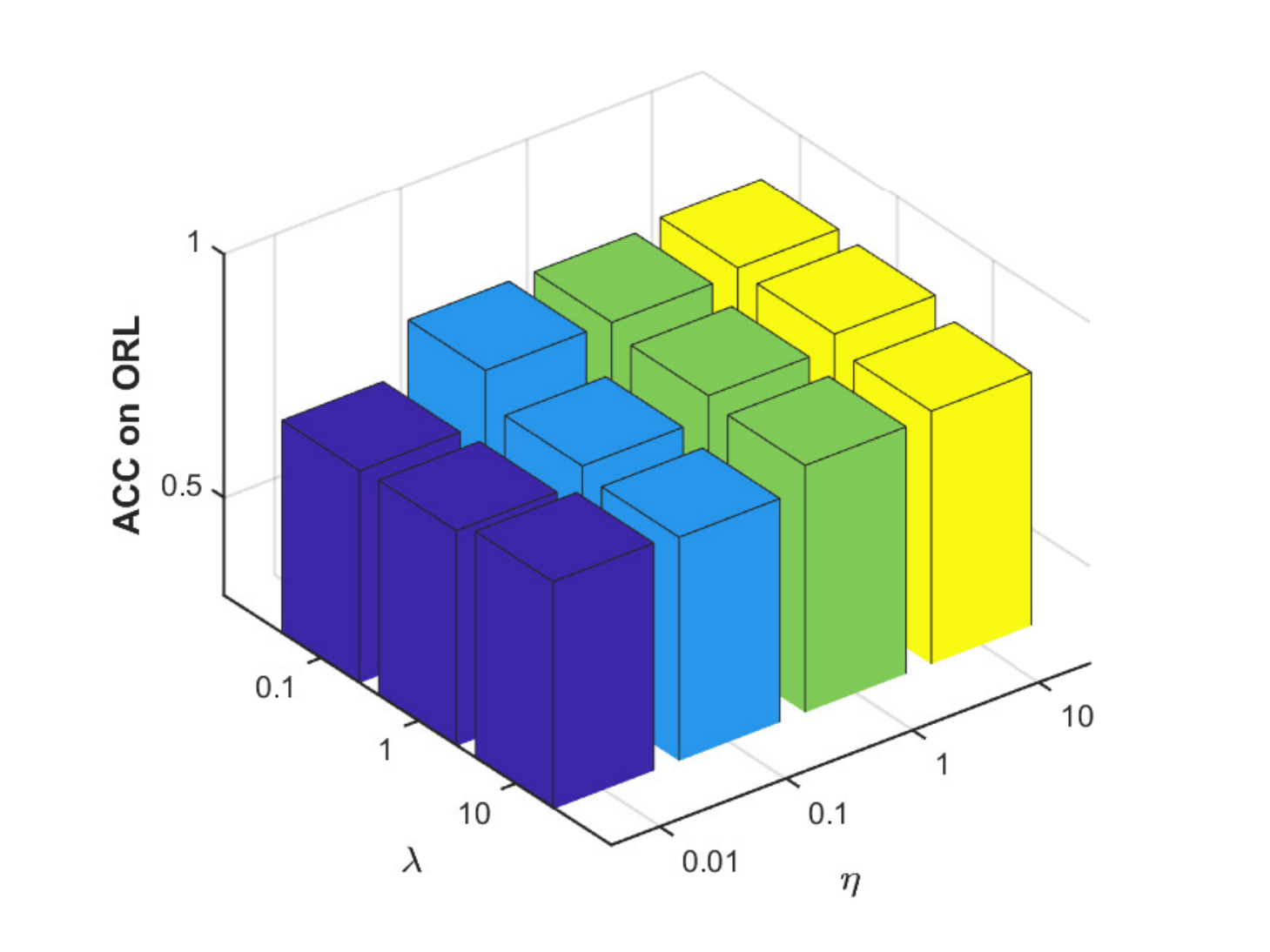}}
\caption{Parameters analysis for FGL-MSC on 100leaves, MSRC-v1 and ORL}
\label{parameters} 
\end{figure*}

Our proposed model has four parameters $\alpha$,$\lambda$,$\eta$, and $m$ that need to be tuned. Among them, $m$ determines the degree of nodes in the learned graph $G$. The setting of $m$ usually depends on the size of the dataset. In our experiments, $m$ is set to 1000 for Hdigit and 10 for the other seven datasets. We also found that for many datasets, the performance tends to be the best when $\alpha$ = $10^{-2}$. Taking 100leaves, MSRC-v1, and ORL for examples, we show the results under different parameters in Fig.\ref{parameters}. In summary, the optimal parameters $(\alpha,\lambda,\eta)$ are set to ($10^{-2}$, 1, 10), ($10^{-2}$, 1, 10), ($10^{-2}$, $10^{-1}$, 10), ($10^{-2}$, 1, $10^{-2}$), ($10^{-2}$, 10, $10^{-2}$), ($10^{-1}$, $10^{-1}$, 10), ($10^{-2}$, 1, $10^{-2}$), ($10^{-2}$, $10^{-1}$, $10^{-2}$) for MSRC-v1, ORL, 100leaves, BBCSport, WebKB, Caltech20, scene15, Hdigit, respectively. It is obvious that our method performs well in a wide range of parameter values.

\subsection{Ablation Study}

To understand the role of the individual modules in FGL-MSC, we propose two variants and conduct experiments on three datasets. The two variants are as follows:

\emph{FGL-Z:} Graph Learning with a two-step updating for $Z^v$ in Eq.(\ref{FGL-MSC}). Firstly, the self-expression matrix $W^v$ is optimized independently by Eq.(\ref{Wp}). Then the objective function Eq.(\ref{FGL-MSC}) becomes:
\begin{equation} \label{FGL-Z}
\begin{aligned}
    &\min_{G,F,A} \sum_{v=1}^t ||W^v-Z^v||_F^2 + \lambda \sum_{i=1}^n ||G_i^T - A_i^T \widetilde{Z}^i||^2_2\\
    & \qquad + \gamma ||G||_F^2 + \eta Tr(F^T L F)\\
    &s.t. \quad G \geq 0,A \geq 0, A_i^T \boldsymbol{1}=1, G\boldsymbol{1}=\boldsymbol{1}, F^T F=I.
\end{aligned}
\end{equation}

In Eq.(\ref{FGL-Z}), the initialization for $Z^v$ is to set $Z^v = W^v$. From a holistic view, the variant FGL-Z is similar to FGL-MSC, except for whether $W^v$ is involved in the joint optimization of the models. 
The learned graph $W^v$ (or $Z^v$) is fused directly in the second step, while the downstream task can no longer have an impact on the update of $W^v$.

\emph{FGL-F:} Graph learning and clustering separately. Specifically, the rank constraint is not considered for the fine-grained graph fusion scheme. The parameter $\eta$ is set to 0, and we implement the spectral clustering method after obtaining $G$ in FGL-F. the objective function of FGL-F can be expressed as follows:
\begin{equation} \label{FGL-F}
\begin{aligned}
    &\min_{G,A} \sum_{v=1}^t \{||X^v-X^v W^v||^{2}_{F} + \alpha||W^v-Z^v||_F^2 \\
    & \qquad + ||W^v||_1\} + \lambda \sum_{i=1}^n ||G_i^T - A_i^T \widetilde{Z}^i||^2_2 +\gamma ||G||_F^2\\
    &s.t. \quad G \geq 0,A \geq 0, A_i^T \boldsymbol{1}=1, G\boldsymbol{1}=\boldsymbol{1},
\end{aligned}
\end{equation}
where $G$ is the input of a classic spectral clustering.

\emph{FGL-Z/F:} Based on the variant FGL-Z, the rank constraint term is also removed. We get the optimal $W^v$ by solving Eq.(\ref{Wp}), and  calculate the unified graph $G$ by Eq.(\ref{FGL-Z/F}). The clustering result can be obtained by spectral clustering on $G$.
\begin{equation} \label{FGL-Z/F}
\begin{aligned}
    &\min_{G,A} \sum_{v=1}^t ||W^v-Z^v||_F^2 + \lambda \sum_{i=1}^n ||G_i^T - A_i^T \widetilde{Z}^i||^2_2\\
    & \qquad + \gamma ||G||_F^2\\
    &s.t. \quad G \geq 0,A \geq 0, A_i^T \boldsymbol{1}=1, G\boldsymbol{1}=\boldsymbol{1}.
\end{aligned}
\end{equation}

\begin{table}
\centering
  \caption{ACC on MSRC-v1, ORL and 100leaves}
  \label{tab:ablation}
  \begin{tabular}{lccc}
    \toprule
    Methods&MSRC-v1&ORL&100leaves\\
    \midrule
    FGL-Z & 0.7338$\pm$0.00 & 0.8375$\pm$0.00 & 0.9060$\pm$0.00 \\
    FGL-F & 0.7493$\pm$0.04 & 0.8358$\pm$0.02 & 0.9109$\pm$0.02\\
    FGL-Z/F & 0.7162$\pm$0.04 & 0.8176$\pm$0.02 & 0.8868$\pm$0.02\\
    FGL-MSC & \textbf{0.7551$\pm$0.00} & \textbf{0.8417$\pm$0.00} & \textbf{0.9522$\pm$0.00}\\
  \bottomrule
\end{tabular}
\end{table}
Table.\ref{tab:ablation} demonstrates that our unified framework containing the graph refinement and rank constraint modules works best under the same dataset and parameters setting. According to the results of FGL-Z/F and FGL-F, the joint optimization for $W^v$ and $Z^v$ outperforms the two-step scheme, but the graph learned by variant FGL-F does not obtain the appropriate cluster structure, with a loss of performance. Comparing the results of FGL-Z/F and FGL-Z, the rank constraint term improves the accuracy of clustering. But the variant FGL-Z can not work well because the graphs generated by the self-expression term do not take full advantage of the multi-view information, thus fitting the clustering task not well enough. 

 \subsection{Effectiveness of Fine-grained Graph Learning}

\begin{figure}
\centering
\vspace{0.1in}
\subfigtopskip=-1pt 
\subfigcapskip=-5pt 
\subfigure[$W^1$]{
	\label{l_1} 
	\includegraphics[width=0.45\linewidth]{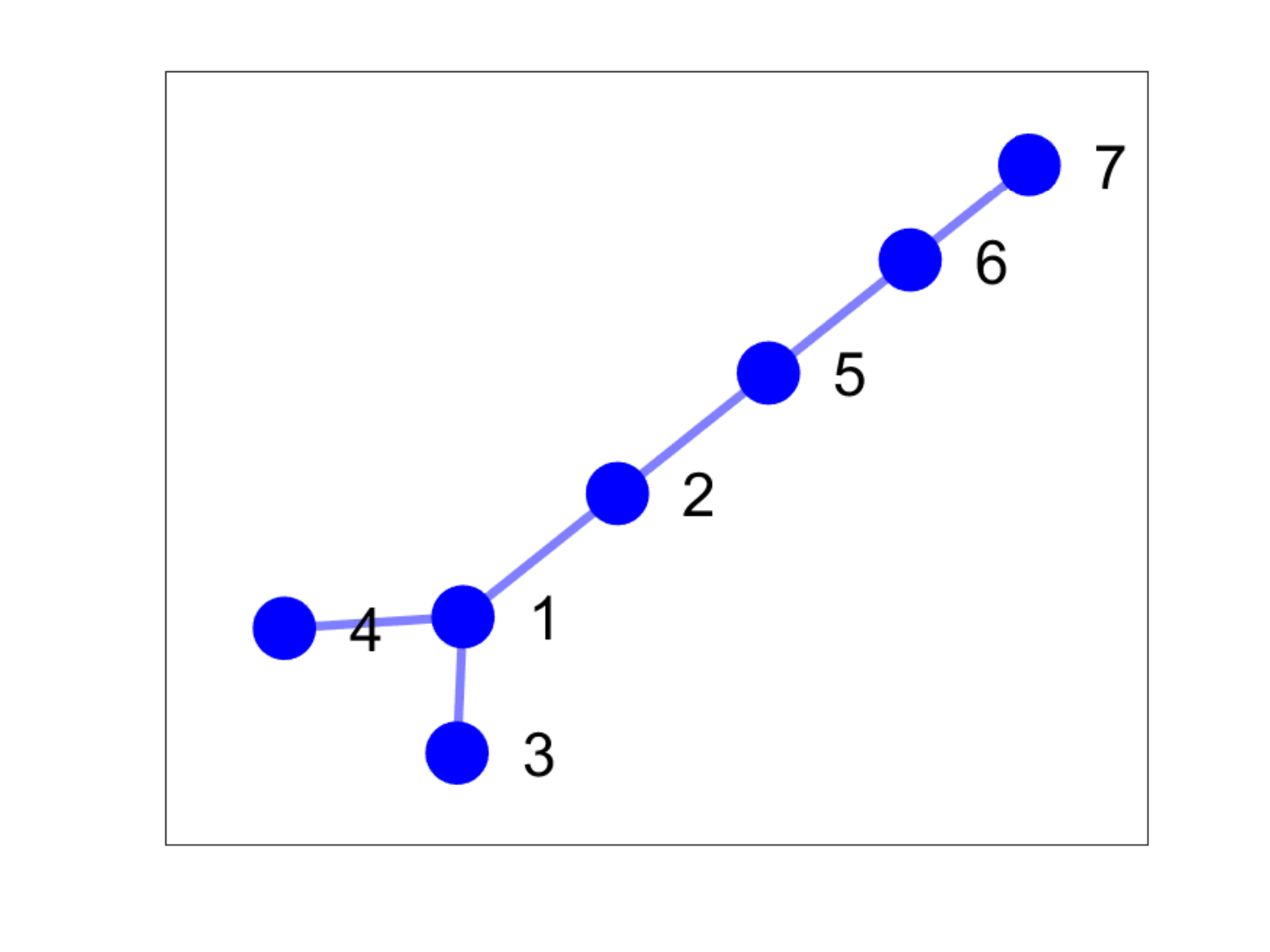}}
\hspace{0.1in}
\subfigure[$W^2$]{
	\label{l_2}
	\includegraphics[width=0.45\linewidth]{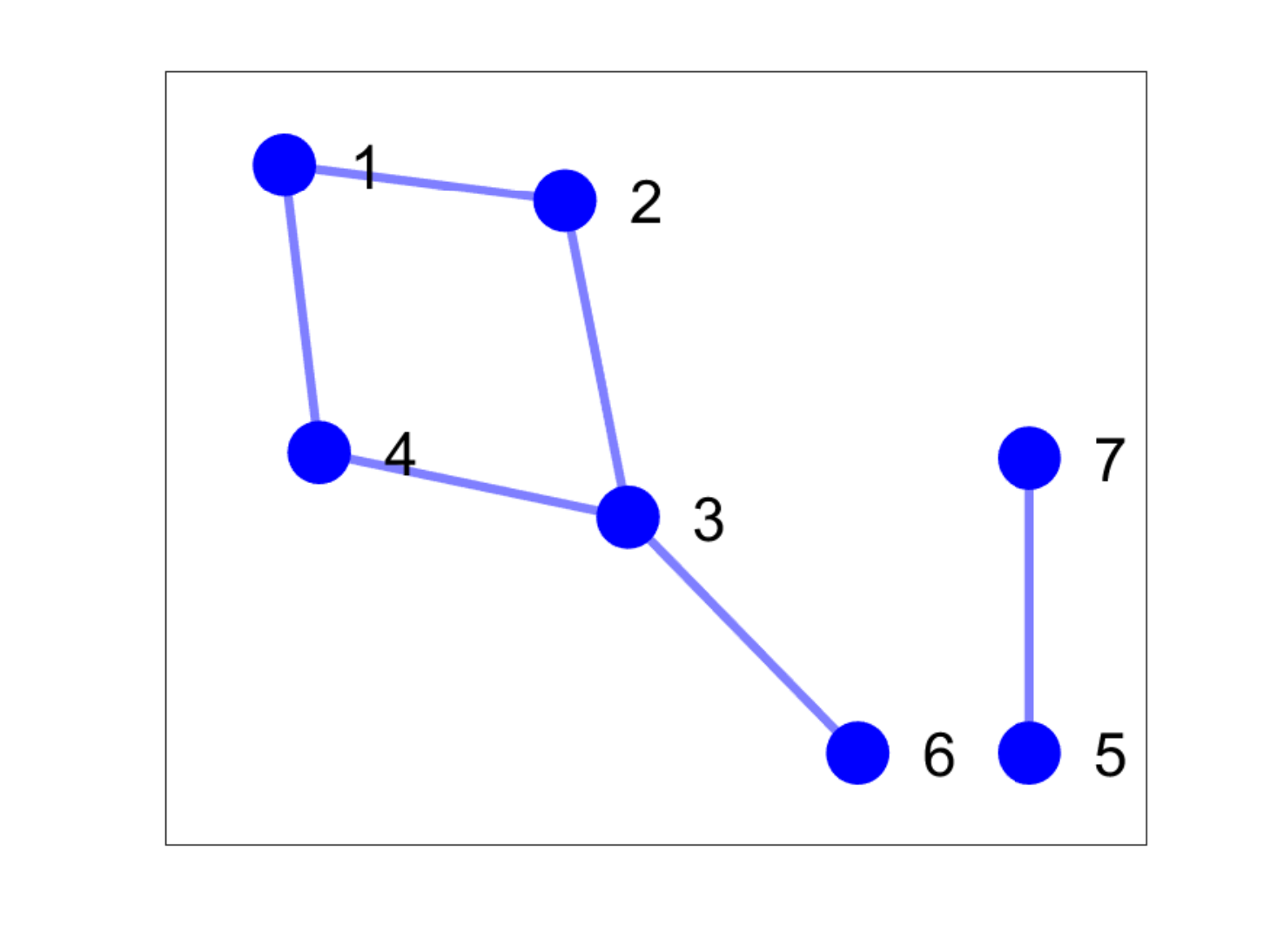}}
\hspace{0.1in}
\subfigure[$W^3$]{
	\label{l_3}
	\includegraphics[width=0.45\linewidth]{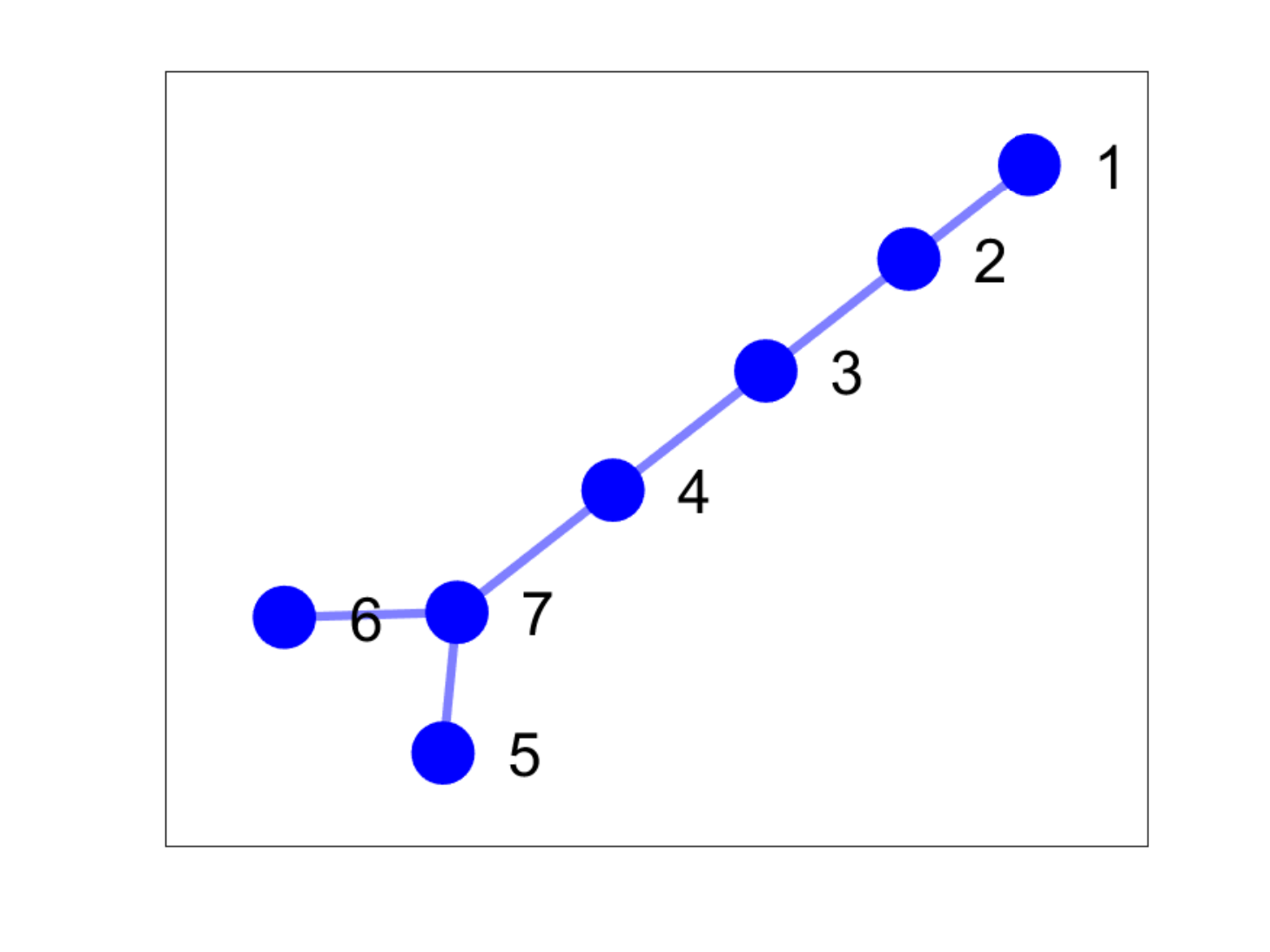}}
 \hspace{0.1in}
\subfigure[$W^4$]{
	\label{l_3}
	\includegraphics[width=0.45\linewidth]{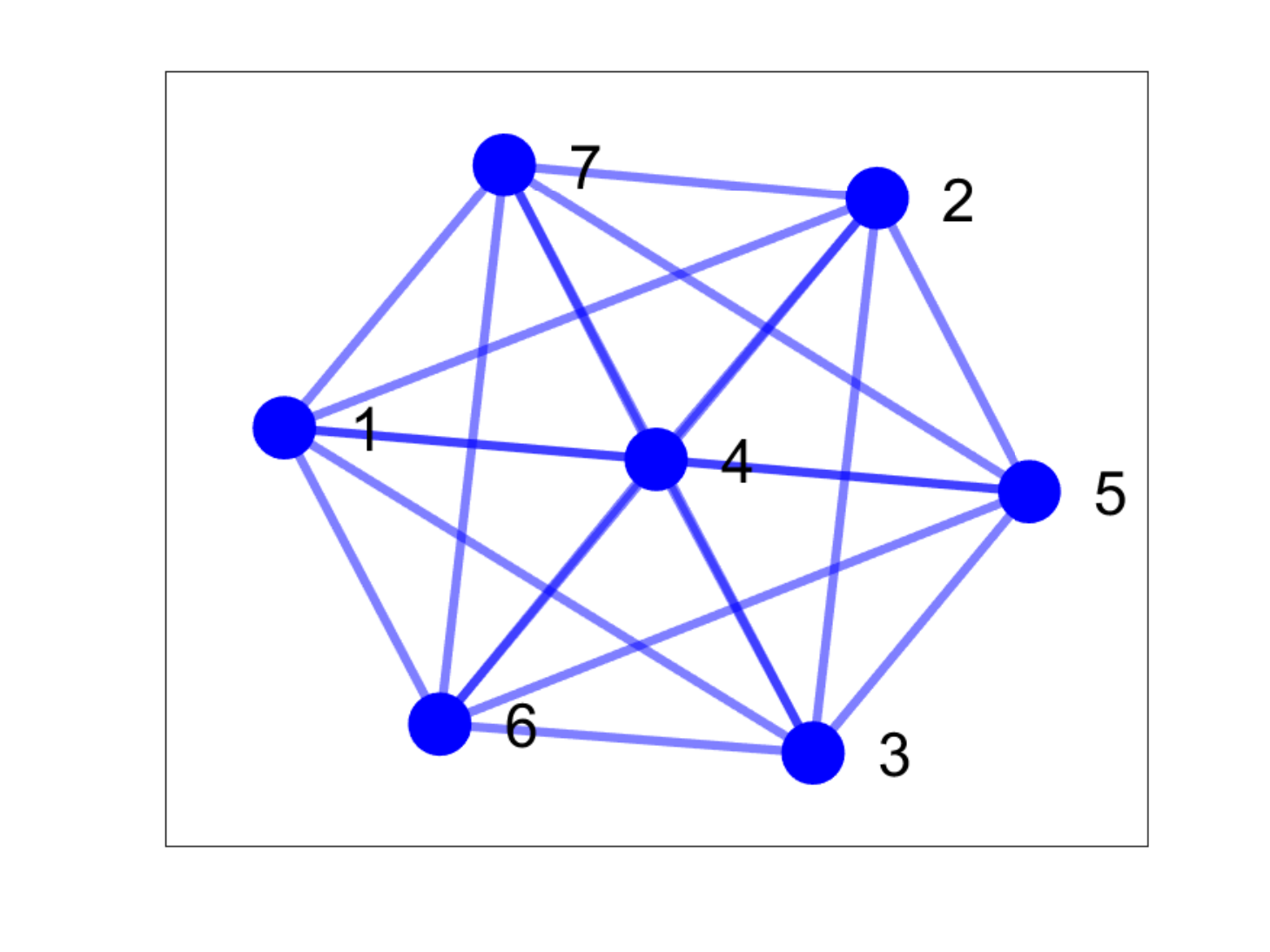}}

\caption{Visualization of the original graphs $W^v$. In the clustering task, nodes 1-4 are in one class and 5-7 are in another.}
\label{graphs} 
\end{figure}

Here we design experiments on synthetic data and three real-world datasets to verify the effectiveness of the fine-grained graph fusion scheme. We introduce a variant of the graph fusion stage, i.e., graph-level fusion instead of fine-grained graph fusion, and the new model is shown in Eq.(\ref{graph-level}). 

\begin{equation} \label{graph-level}
\begin{aligned}
    &\min_{G,F,a} \sum_{v=1}^t \{||X^v-X^v W^v||^{2}_{F} + \alpha||W^v-Z^v||_F^2 \\
    & \qquad + ||W^v||_1\} + \lambda  ||G - \sum_{v=1}^t{a_v Z^v}||^2_F +\gamma ||G||_F^2\\
    & \qquad + \eta Tr(F^T L F)\\
    &s.t. \quad G \geq 0,a \geq 0, a^T \boldsymbol{1}=1, G\boldsymbol{1}=\boldsymbol{1}, F^T F=I,
\end{aligned}
\end{equation}
the optimization of Eq.(\ref{graph-level}) differs from that of FGL-MSC in the update of the fusion weight $a$.

\begin{equation} \label{new a}
\begin{aligned}
    &\min_{a} ||G - \sum_{v=1}^t{a_v Z^v}||^2_F\\
    &s.t. \quad a \geq 0, a^T \boldsymbol{1}=1,
\end{aligned}
\end{equation}
\cite{AMUSE} provides an efficient algorithm for solving Eq.(\ref{new a}), and we substitute the weight $a$ derived from this algorithm into the update process for the remaining variables to obtain the model results for the graph-level fusion.

\begin{figure}
\centering
\vspace{0.1in}
\subfigtopskip=-1pt 
\subfigcapskip=-5pt 
\subfigure[Graph-level Fusion]{
	\label{o_1} 
	\includegraphics[width=0.45\linewidth]{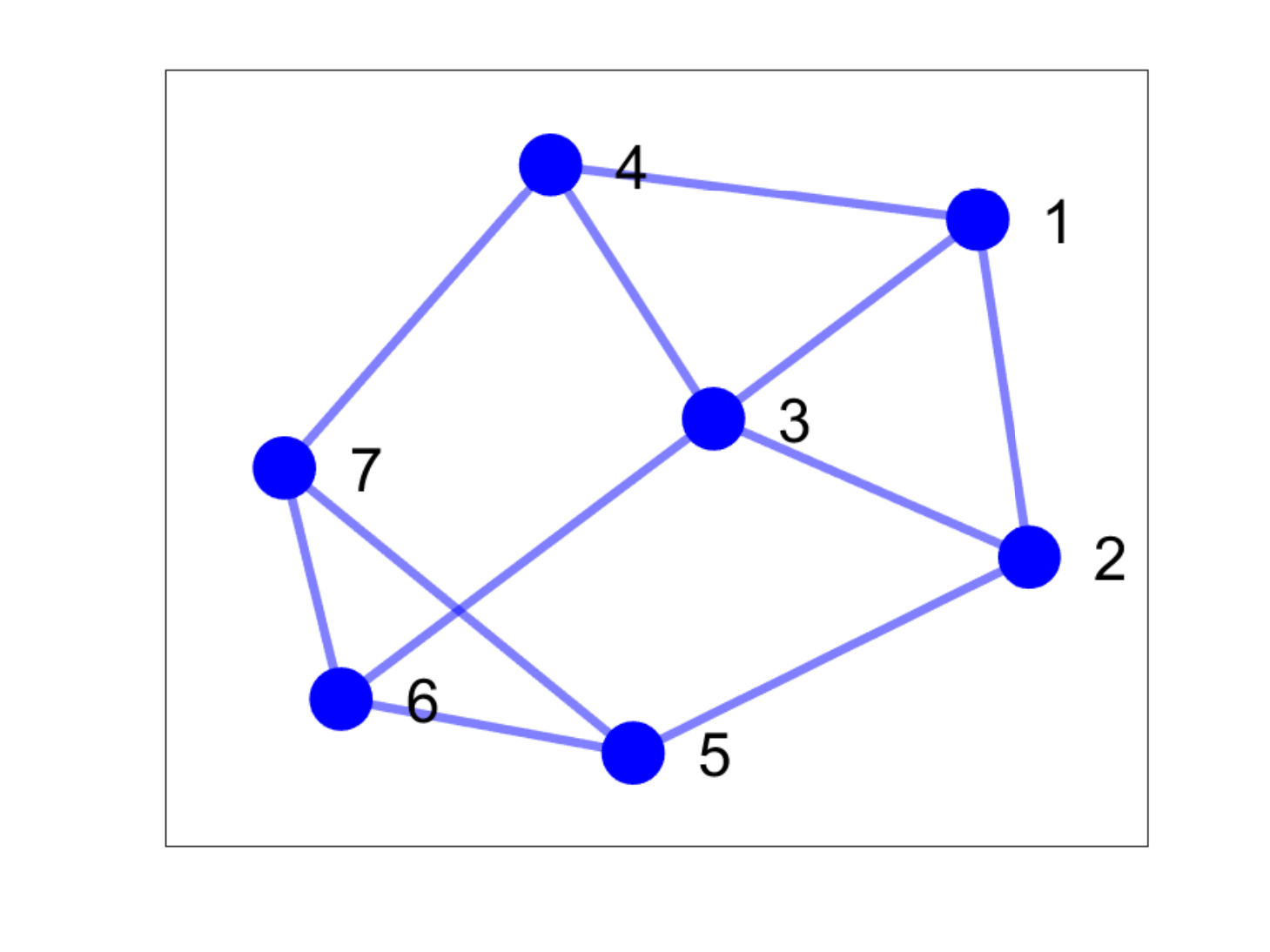}}
\hspace{0.1in}
\subfigure[Fine-grained Fusion]{
	\label{o_2}
	\includegraphics[width=0.45\linewidth]{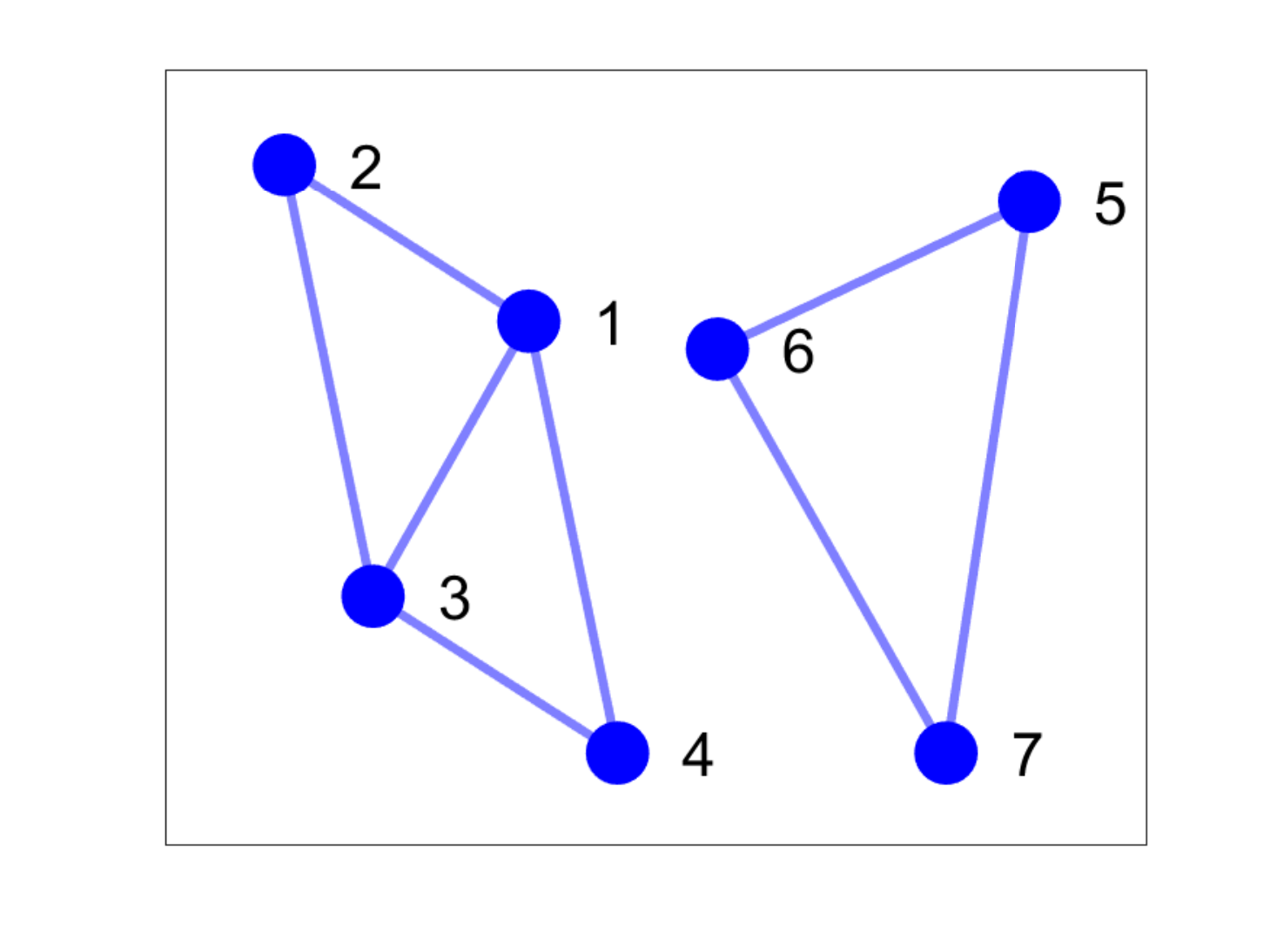}}
\caption{The unified graphs generated by the graph-level fusion method and FGL-MSC.}
\label{graph fusion} 
\end{figure}

\begin{figure*}
\centering
\vspace{0.1in}
\subfigtopskip=-1pt 
\subfigcapskip=-5pt 
\subfigure[MSRC-v1]{
	\label{o_1} 
	\includegraphics[width=0.3\linewidth]{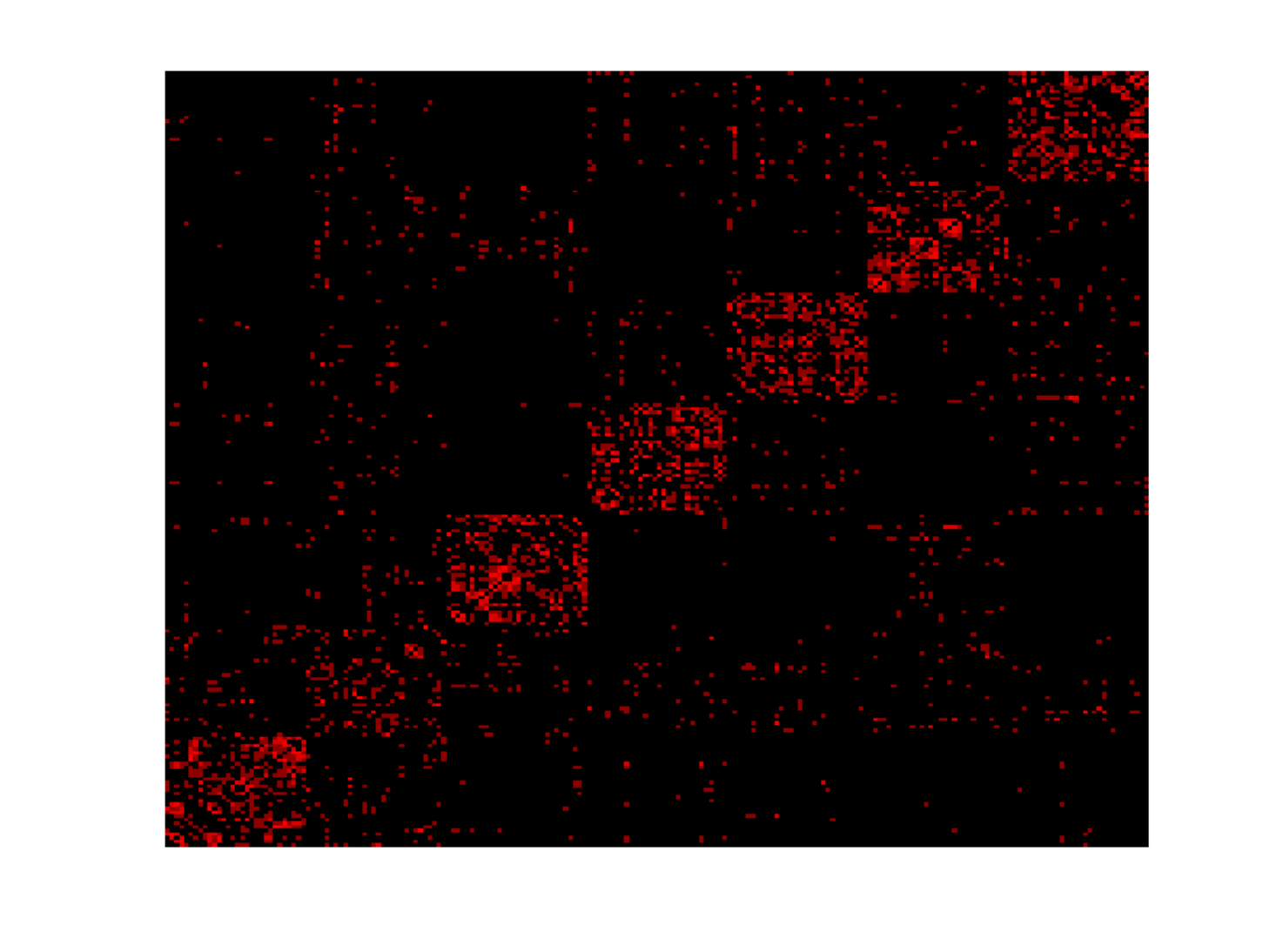}}
\hspace{0.1in}
\subfigure[Caltech20]{
	\label{o_2}
	\includegraphics[width=0.3\linewidth]{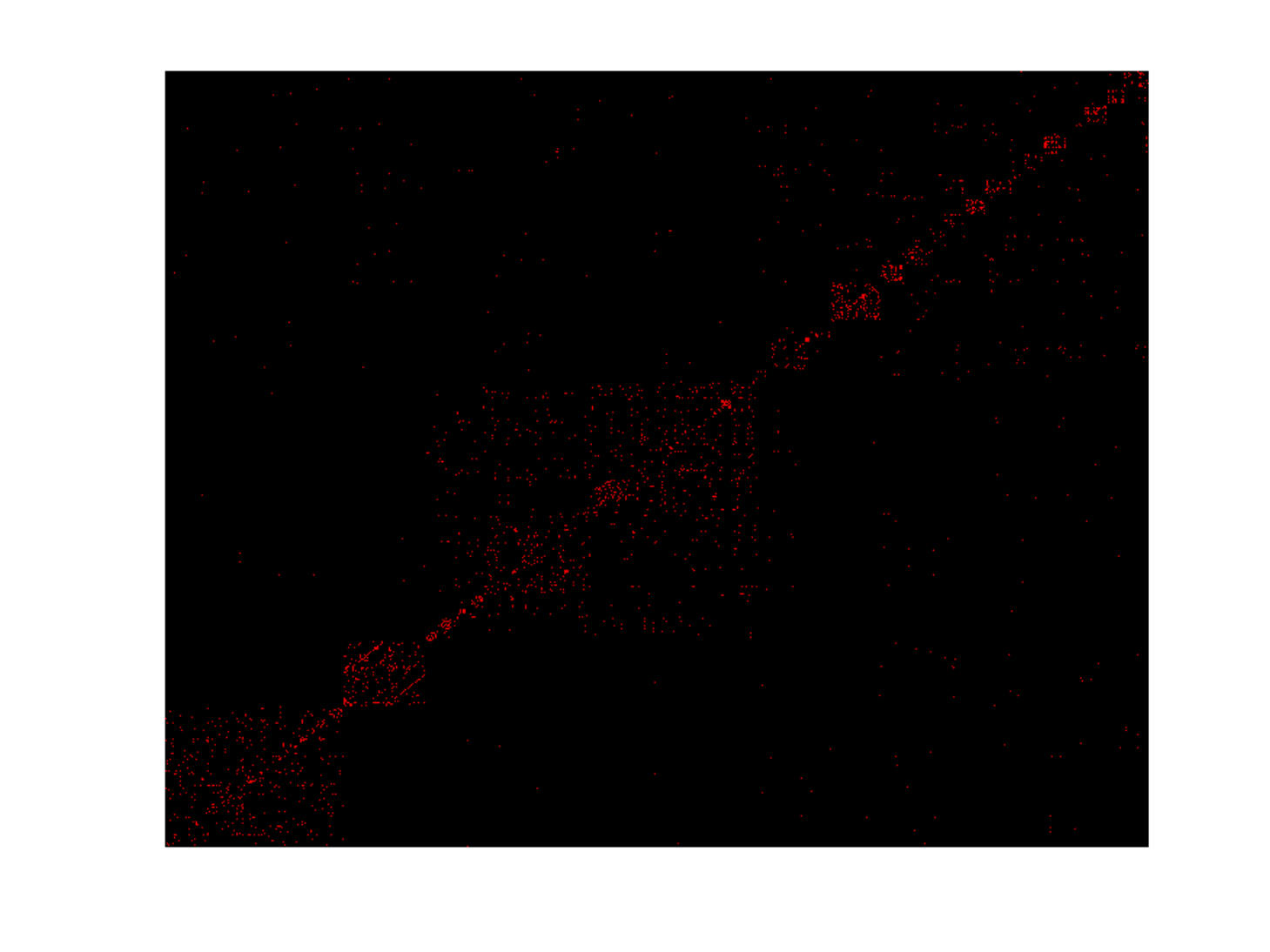}}
\hspace{0.1in}
\subfigure[Hdigit]{
	\label{o_2}
	\includegraphics[width=0.3\linewidth]{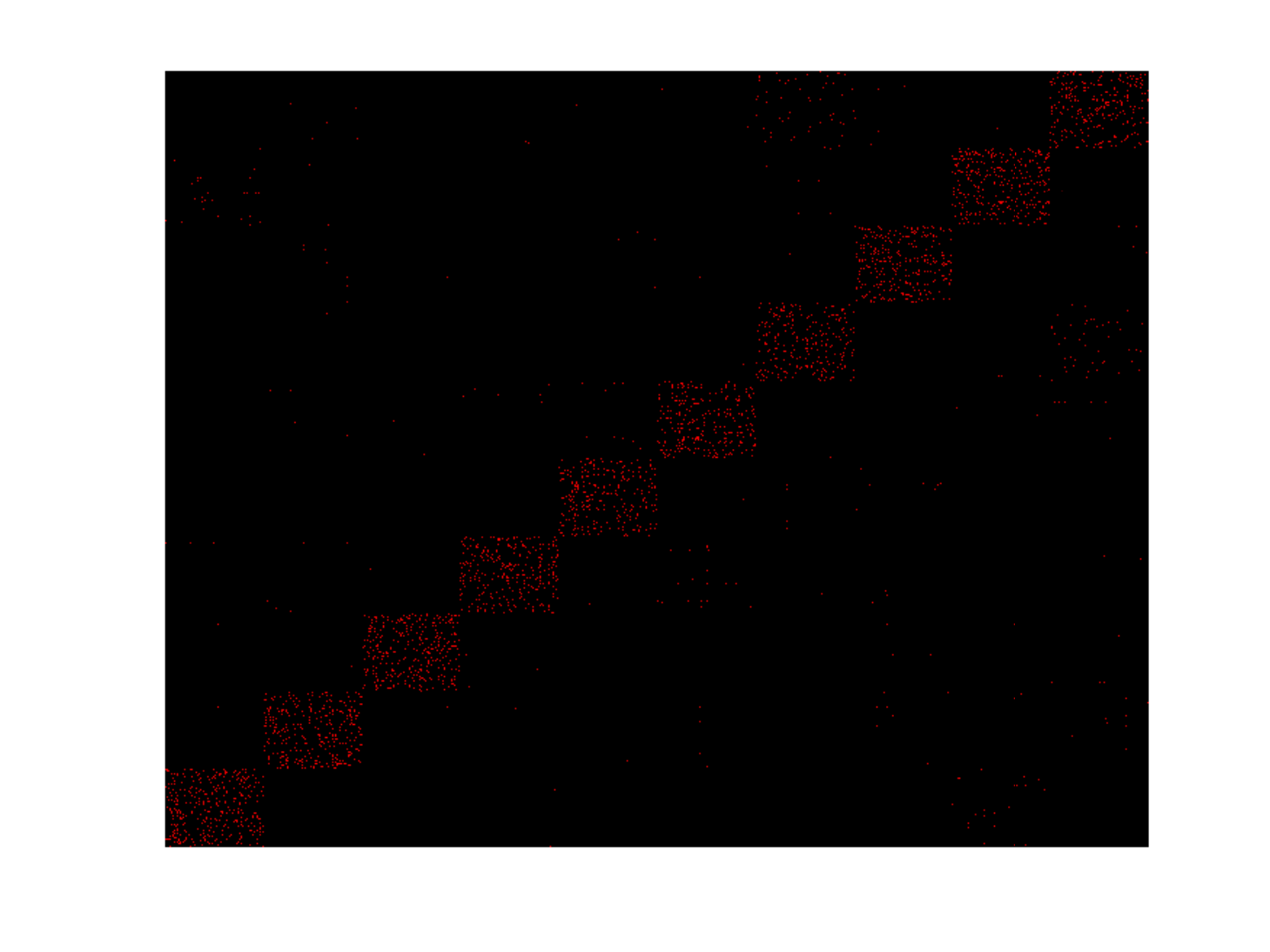}}
\caption{Visualization of the unified graphs generated by FGL-MSC on three datasets.}
\label{graphs in 3} 
\end{figure*}

First, We make a toy dataset containing 7 nodes with 4 sets of their feature views. To visualize clearly, we demonstrate each original graph $W^v$ generated by Eq.(\ref{CAN}) and Eq.(\ref{solveW}), as shown in Fig.(\ref{graphs}). Then we conduct the graph-level fusion method and FGL-MSC to obtain their unified graphs separately. The unified graphs are shown in Fig.(\ref{graph fusion}). Obviously, there are still redundant edges between two clusters in the unified graph obtained by the graph-level fusion method, which will have an impact on the clustering performance. And the fine-grained graph fusion scheme splits two clusters successfully. Therefore, our proposed model has the ability to filter local structures and learn potential clustering representations.
Then we visualize the learned unified graphs by FGL-MSC on MSRC-v1, Caltech20, and Hdigit dataset in Fig.(\ref{graphs in 3}). It is observed that the number of connected subgraphs on each unified graph is approximated by the number of clusters. Also, the unified graphs are sparse with few redundant edges between clusters. These experimental results can demonstrate the effectiveness of the fine-grained graph fusion scheme and explain the good performance achieved by FGL-MSC.

\subsection{Convergence Analysis}

\begin{figure}
\centering
\vspace{0in}
\subfigtopskip=0pt 
\subfigcapskip=-5pt 
\subfigure[ORL]{
	\label{conv1} 
	\includegraphics[width=0.45\linewidth]{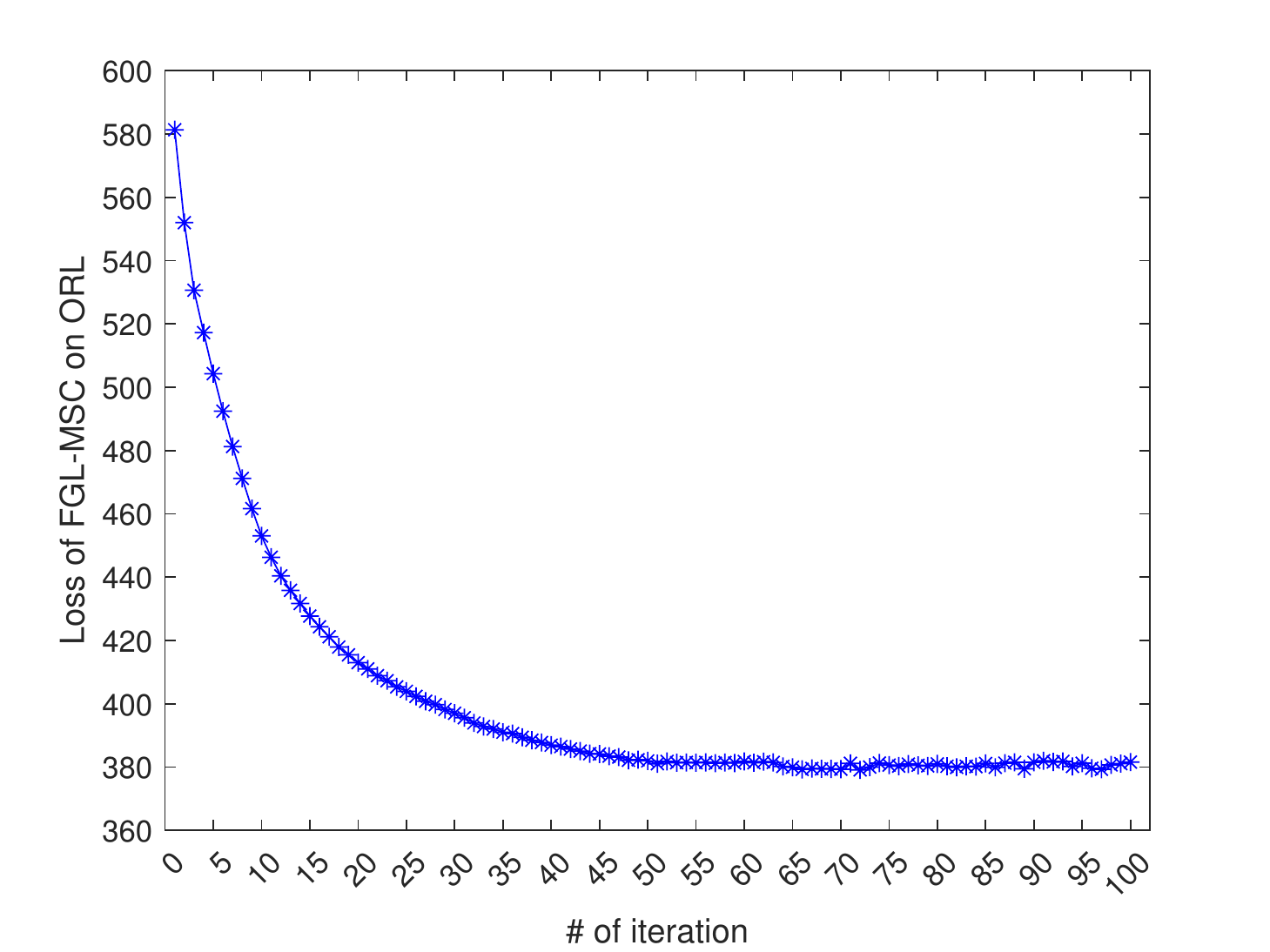}}
\hspace{0.1in}
\subfigure[Caltech20]{
	\label{conv2}
	\includegraphics[width=0.45\linewidth]{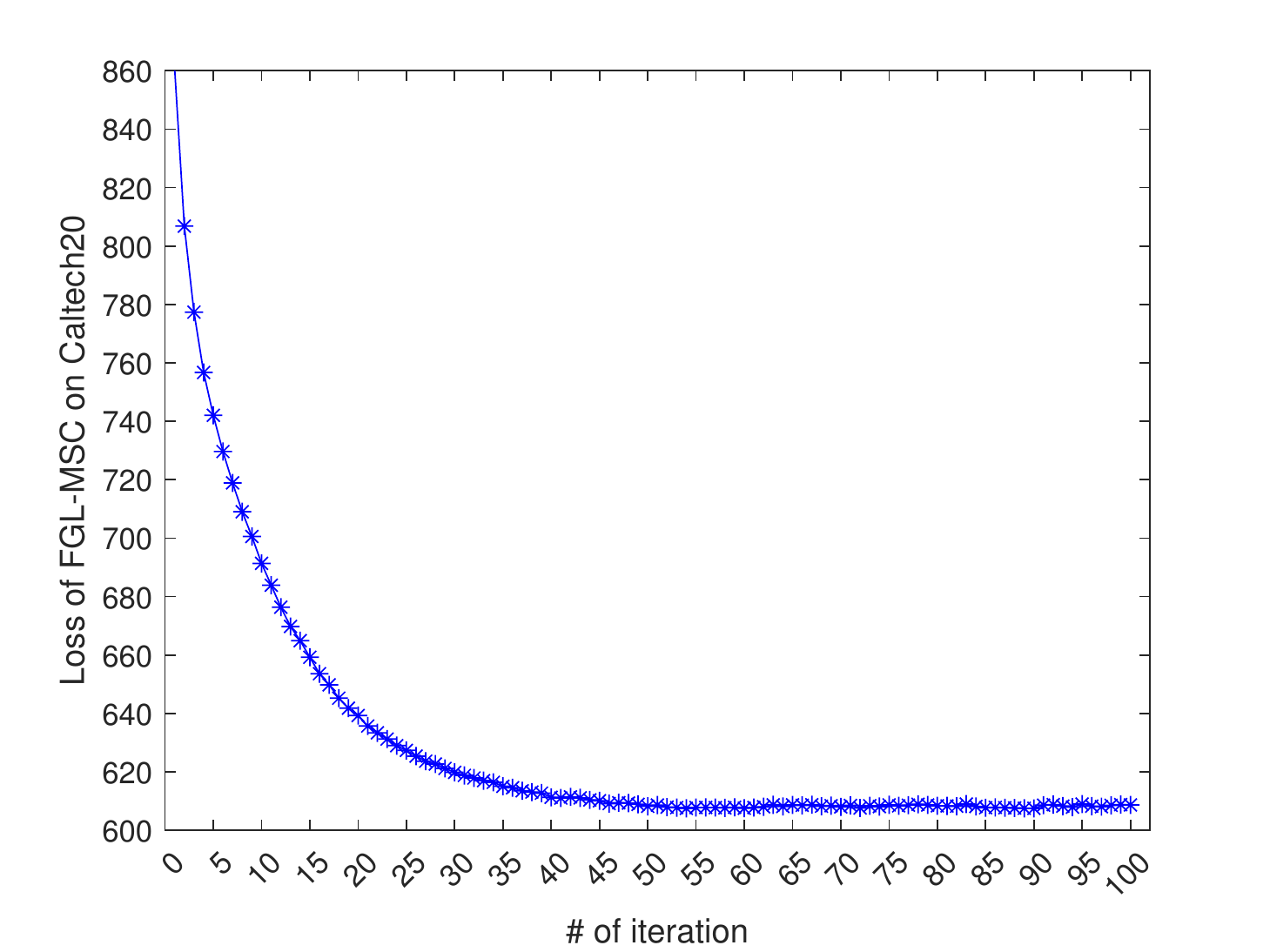}}
\caption{The convergence curves of FGL-MSC on ORL and Caltech20.}
\label{convergence} 
\end{figure}

We investigate the convergence analysis of our model by representing the loss value calculated by Eq.(\ref{FGL-MSC}) with the increasing iteration. The results on ORL and Caltech20
are reported in Fig.\ref{convergence}. It is clear that the loss value is decreasing stably during the optimization. When the iteration runs to 20 rounds, the rates of change are significantly lower. When iterating to 50 rounds, the loss values plateau. The iteration set in our experiments is 10 for computational efficiency considerations. But the convergence results tell us that FGL-MSC could achieve a better performance than the above experiments. Generally speaking, our model has good convergence and achieves cheerful performance after a few iterations.

\section{Conclusion}
 
In this paper, we propose a new fine-grained graph learning framework for multi-view subspace clustering. To avoid the deterioration of the learning process from incorrect neighbors and redundant structures and to elevate the clustering performance, we employ a fine-grained graph fusion method, which is incorporated into the joint framework. This framework generates the fine-grained weights, the unified graph, and the clustering representation simultaneously. An effective optimization strategy with guaranteed convergence was provided to solve the joint objective function. Then, Extensive experiments on some benchmark datasets demonstrated the effectiveness of our method for multi-view subspace clustering.




%





\ifCLASSOPTIONcaptionsoff
  \newpage
\fi





\bibliographystyle{IEEEtran}
\bibliography{IEEEabrv,Bibliography}
%

\begin{IEEEbiography}[{\includegraphics[width=1in,height=1.25in,clip,keepaspectratio]{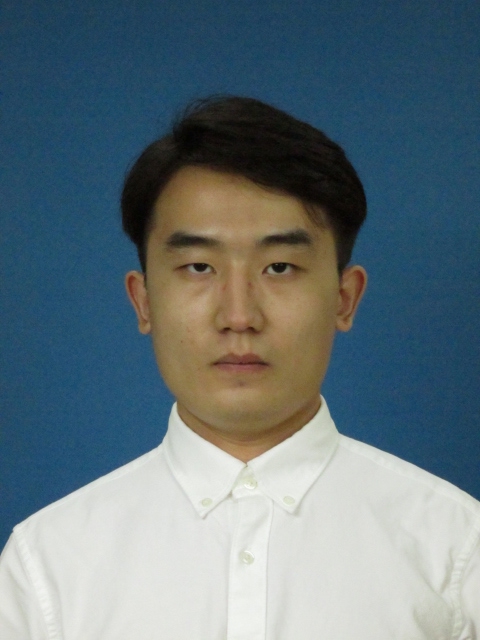}}]{Yidi Wang}
received the B.E. degree in software engineering from Jiangxi University of Science and Technology, Nanchang, China, in 2020, and the master's degree from Huazhong University of Science and Technology, Wuhan, China, in 2023, respectively. His current research interest includes graph machine learning. 
\end{IEEEbiography}
\begin{IEEEbiography}[{\includegraphics[width=1in,height=1.25in,clip,keepaspectratio]{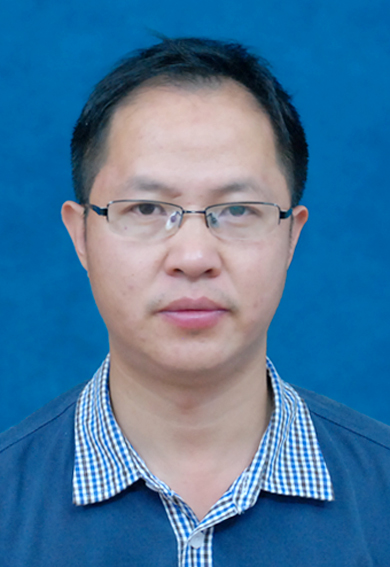}}]{Xiaobing Pei}
received a Ph.D. degree in computer software and theory from Huazhong University of Science and Technology, Wuhan, China, in 2006. From 2000 to 2003, he was a Research Engineer and a Software Developer with the Research and Development Department, Huawei Technologies Company Ltd., Shenzhen, China. He is currently a professor with the School of Software, Huazhong University of Science and Technology. His current research interests include machine learning, data mining and software engineering.
\end{IEEEbiography}
\begin{IEEEbiography}[{\includegraphics[width=1in,height=1.25in,clip,keepaspectratio]{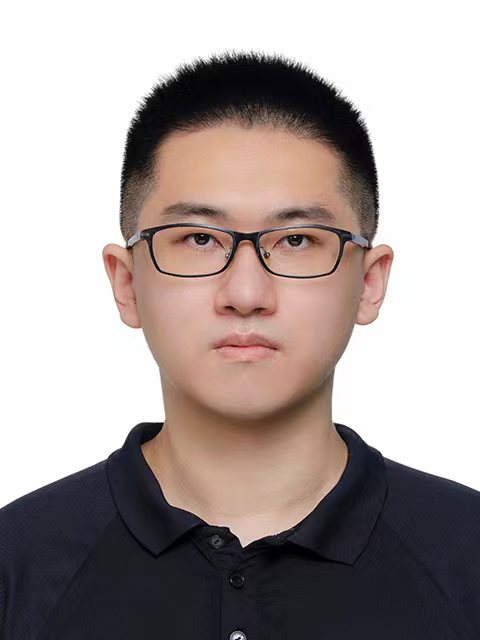}}]{Haoxi Zhan}
received the B.A. degree from Hamp shire College, USA, in 2016, and the master's degree from Huazhong University of Science and Technology, Wuhan, China, in 2022, respectively. He is currently working toward the Ph.D. degree with the Faculty of Electronic Information and Electrical Engineering, Dalian University of Technology, Dalian, China. His research interests include machine learning, graph neural networks, and cognitive science.
\end{IEEEbiography}




\vfill


\end{document}